\documentclass[10pt,journal,compsoc]{IEEEtran}

\usepackage{xspace}
\usepackage{times}
\usepackage{epsfig}
\usepackage{graphicx}
\usepackage{amsmath}
\usepackage{amssymb}
\usepackage{booktabs}  
\usepackage{textcomp}  
\usepackage[utf8]{inputenc}
\usepackage[T1]{fontenc}
\usepackage{epigraph}
\usepackage{csquotes}
\usepackage{bm}
\usepackage[pagebackref=true,breaklinks=true,letterpaper=true,colorlinks=false,bookmarks=false]{hyperref}

\newcommand{\etal}{\textit{et al.}}

\newcommand{\fid}{Fréchet Inception Distance\xspace}
\newcommand{\fids}{Fréchet Inception Distances\xspace}

\newcommand{\warped}{warped patches}        
\newcommand{\dewarped}{dewarped patches}    

\newcommand{\srcimage}{\mathbf{x}_{s}}                         
\newcommand{\genimage}{\tilde{\mathbf{x}}}                     
\newcommand{\srcgenimage}{\tilde{\mathbf{x}}_{s}}              
\newcommand{\dstgenimage}{\tilde{\mathbf{x}}_{d}}              

\newcommand{\srcviewpoint}{\mathbf{V}_{s}}                     
\newcommand{\dstviewpoint}{\mathbf{V}_{d}}                     

\newcommand{\genericviewpoint}{\mathbf{V}}                         

\newcommand{\srcplset}{\bm{\mathcal{P}}_{s}}                         
\newcommand{\srcplsetunwarp}{\bm{\tilde{\mathcal{P}}_{s}}}           
\newcommand{\dstplset}{\bm{\mathcal{P}}_{d}}                         

\newcommand{\srcpl}{\mathbf{p}_{s}}                            
\newcommand{\srcplunwarp}{\tilde{\mathbf{p}}_{s}}              
\newcommand{\dstpl}{\mathbf{p}_{d}}                            

\newcommand{\srccrop}{c_s}                                     

\newcommand{\srckpoints}{\bm{\mathcal{K}}_{s}}                       
\newcommand{\dstkpoints}{\bm{\mathcal{K}}_{d}}                       

\newcommand{\sketches}{s_{2.5D}}                         
\newcommand{\sketchspace}{{\rm I\!R}^{H\!\times W\!\times 3}}   

\newcommand{\cadset}{\mathcal{C}}                              
\newcommand{\cadith}{C^{(i)}}                                  

\newcommand{\ith}{^{(i)}}                                   

\newcommand{\imagespace}{{\rm I\!R}^{H\!\times W\!\times 3}}   
\newcommand{\kpointspace}{{\rm I\!R}^{|\srckpoints|\!\times 2}}            
\newcommand{\vpointspace}{{\rm I\!R}^{4\!\times 4}}            
\newcommand{\meshspace}{{\rm I\!R}^{faces \times 3 \times 3}}            

\newcommand{\given}{\,\,|\,\,}

\newcommand{\vunet}{\textit{VUnet}\xspace}
\newcommand{\von}{\textit{VON}\xspace}
\newcommand{\vonfinetuned}{\textit{VON}$_{FT}$\xspace}

\newcommand{\pascal}{Pascal3D+\xspace}

\newcommand{\plain}{\textit{plain}\xspace}
\newcommand{\textured}{\textit{textured}\xspace}

\newcommand{\icn}{ICN\xspace}

\newcommand{\MvToNv}{MV2NV\xspace}
\newcommand{\MvThreeD}{MV3D\xspace}
\newcommand{\tvsn}{TVSN\xspace}

\begin{document}

\title{Warp and Learn: Novel Views Generation for Vehicles and Other Objects}

\author{
Andrea Palazzi$^\ast$,
Luca Bergamini$^\ast$,
Simone Calderara,
Rita Cucchiara
\thanks{All authors are with University of Modena and Reggio Emilia, Italy.\protect\\
E-mail: name.surname@unimore.it\protect\\
$^\ast$ indicates equal contribution.}
}

\markboth{IEEE TRANSACTIONS ON PATTERN ANALYSIS AND MACHINE INTELLIGENCE}%
{Palazzi \MakeLowercase{\textit{et al.}}: Warp and Learn: Novel Views Generation for Vehicles and Other Objects}


\IEEEtitleabstractindextext{
    \begin{abstract}
   In this work we introduce a new self-supervised, semi-parametric approach for synthesizing novel views of a vehicle starting from a single monocular image. Differently from \textit{parametric} (i.e. entirely learning-based) methods, we show how \textit{a-priori} geometric knowledge about the object and the 3D world can be successfully integrated into a deep learning based image generation framework. As this geometric component is not learnt, we call our approach \textit{semi-parametric}.
   In particular, we exploit man-made object symmetry and piece-wise planarity to integrate rich \textit{a-priori} visual information into the novel viewpoint synthesis process. An Image Completion Network (\icn) is then trained to generate a realistic image starting from this geometric guidance. 
   This careful blend between parametric and non-parametric components allows us to i) operate in a real-world scenario, ii) preserve high-frequency visual information such as textures, iii) handle truly arbitrary 3D roto-translations of the input and iv) perform shape transfer to completely different 3D models. Eventually, we show that our approach can be easily 
   complemented with synthetic data and extended to other rigid objects with completely different topology, even in presence of concave structures and holes (e.g. chairs). 
   A comprehensive experimental analysis against state-of-the-art competitors shows the efficacy of our method both from a quantitative and a perceptive point of view.\\ \\
   Supplementary material, animated results, code and data are available at:\\
   \href{https://github.com/ndrplz/semiparametric}{https://github.com/ndrplz/semiparametric}
\end{abstract}
    \begin{IEEEkeywords}
    Vehicle novel view generation, Novel viewpoint synthesis, Texture transfer, Shape transfer, Semi-parametric.
    \end{IEEEkeywords}
}

\maketitle
\IEEEdisplaynontitleabstractindextext
\IEEEpeerreviewmaketitle

\maketitle

\section{Introduction}
\noindent \IEEEPARstart{H}{ow} would you see an object from another point of view? Given a single view of an object in the world, predicting how it would look like from arbitrarily different viewpoints is definitely non-trivial for both humans and machines.
Still, people with a good \textit{visual-spatial intelligence}~\cite{gardner2011frames} can easily imagine objects' rotation, zoom and translation shifts, especially if objects have a well known shape and feature some degree of symmetry. Indeed, humans have been shown to perform mental transformations for decision-taking about their surrounding environment~\cite{shepard1971mental,bundesen1975visual}.\\
Machines are still far from this level of intelligence.
%
\begin{figure*}[t]
    \centering
    \includegraphics[width=\textwidth]{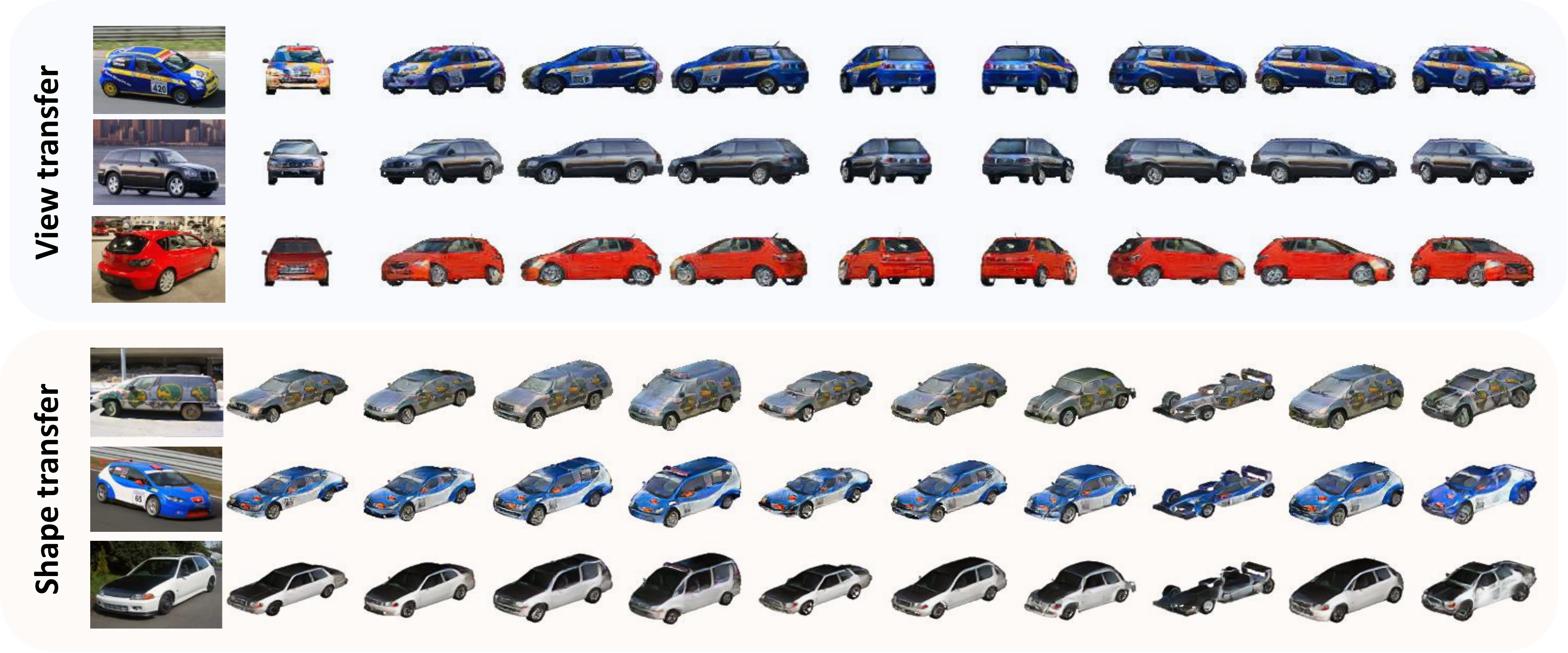}
    \caption{We propose a semi-parametric framework to generate realistic novel views of a vehicle and / or to transfer its appearance to different models.}
    \label{fig:teaser}
\end{figure*}
%
Still, powerful \textit{parametric} (i.e. entirely learning-based) deep learning models~\cite{kingma2013auto, goodfellow2014generative} made it possible to frame the generation of novel viewpoints as a \textit{conditioned image synthesis} problem. 
However, this is an holistic approach that under-exploits the fact that man-made objects 3D models are roughly distributed according to few prototypes (e.g. sedan, VAN, pick-up, truck etc. for vehicles). Up to now, the generation is constrained to be visually plausible with almost no geometric support from prototypes' shapes \cite{tatarchenko2016multi,park2017transformation}.
Furthermore, even though generated images may look realistic \textit{per se}, fine-grained visual appearance characterizing the particular object instance (e.g. texture) is often lost due to its high frequency which hardly survives being encoded through a deep network \cite{tatarchenko2016multi,zhu2018visual,esser2018variational}.
Eventually, most methods are supervised on the target image, which is seldom available for real data. Also, vast amount of data are required for the network to generalize to arbitrary transformations (i.e. a sufficient number of images for every possible viewpoint). 
This constrains many methods to be trained solely on synthetic data and to be restricted to a discrete set of viewpoints: in fact, most recent works only handle a small set of transformations (e.g. fixed radius rotations)~\cite{yang2015weakly,tatarchenko2016multi,zhou2016view,park2017transformation}.\\
%
%
\indent At the same time, an independent line of research has shown that a \textit{non-parametric} approach can be a viable path for photorealism, as also pointed out by Qi~\etal~\cite{qi2018semi}. For instance, new images can be generated by collaging~\cite{hays2007scene,lalonde2007photo,chen2009sketch2photo,isola2013scene} or by leveraging multiple photographs to synthesize novel views via image-based rendering~\cite{chaurasia2013depth,ortiz2015bayesian,hedman2016scalable,ortiz2016automatic}. Still, these methods require a large amount of data at test time: entire image banks for collaging, multiple photographs and depth data for image-based rendering.\\
\indent In this work we propose a new approach - inherently \textit{semi-parametric} - being based on both learning and geometry, self-supervised and efficient to be used in real-time. By taking the best from both worlds, we exploit geometric constraints to roughly sketch the target shape of the object and its textures while still relying on deep view synthesis to refine the generated view.  
%
The rationale behind this work is that many man-made objects adhere to a-priori geometric rules: vehicles in particular, exhibit a symmetric, piece-wise planar structure. Therefore, those properties may be exploited to approximately represent them by a small set of piece-wise planar patches, which can be warped almost exactly from source to destination viewpoint via a symmetry-aware homography transformation.
Although these warped patches provide a rich hint about the visual content of the target viewpoint, they are far from being useful on their own. Thus, a fully-convolutional network is seeded with these patches along with a 2.5D CAD-rendered sketch to be used as guidance; it is then trained in a self-supervised manner to discriminate which part of the image must be completed or in-painted for the result to look realistic (see Fig.~\ref{fig:architecture}).\\
We tested our solution in particular in vehicle generation due to their ubiquity in urban scene understanding applications~\cite{yang2018denseaspp,teichmann2018multinet,feng2018towards,shen2017learning,marin2018unsupervised,liu2016large}.
Moreover, to highlight that a decomposition in planar patches holds for different types of rigid objects, we evaluate our semi-parametric framework on both convex objects (\textit{vehicles}) and concave ones (\textit{chairs}). We leave as future work the analysis of a broader set of object categories.\\

In summary, our main contributions follow:
\begin{itemize}
    \item We propose an original formulation of the problem of object novel viewpoint synthesis in a semi-parametric setting. Loose geometrical assumptions about the object shape provide rich hints about its appearance (\textit{non-parametric}); this information guides a fully-convolutional network (\textit{parametric}) in the generation process.
    \item We design our model to be trainable on existing datasets for 3D object detection in a \textit{self-supervised} manner, with no need for paired source/target viewpoint images. Furthermore, we leverage 2D keypoints for real-world images where foreground segmentation is not provided.
    \item We demonstrate how our method excels in preserving visual details (e.g. texture) and in performing realistic shape transfer to completely different 3D models, while still being resilient to a much wider range of 3D transformations than competitors.
\end{itemize}
Our method can be employed to generate realistic novel views of an object from an arbitrary zoom, viewpoint and distance, as depicted in Figures~\ref{fig:teaser},~\ref{fig:rotation_results},~\ref{fig:competitors},~\ref{fig:cameras},~\ref{fig:synth_compare},~\ref{fig:chairs_comparison},~\ref{fig:armchair_backflip}. This could enable the generation of novel views from arbitrary virtual cameras in an urban scene, with applications in surveillance, vehicle re-identification and forensics. Also, our approach allows a disentangled editing of object shape and appearance (i.e. shape can be changed while preserving appearance or the other way around). This enables applications in interactive 3D manipulation and design, as well as data augmentation (Fig.~\ref{fig:pascal_aug}).
%
%
\\ \\
\noindent A thorough experimental analysis is conducted comparing our proposal with state-of-the-art methods, considering both the quantitative and the perceptual point of view.

\section{Related Work}
\noindent \textbf{View synthesis}
In just few years, the widespread adoption of deep generative models~\cite{kingma2013auto, goodfellow2014generative} has led to astounding results in different areas of image synthesis~\cite{radford2015unsupervised,arjovsky2017wasserstein,zhang2017stackgan,karras2017progressive,wang2018high,yu2018generative}. In this scenario, conditional GANs~\cite{mirza2014conditional} have been demonstrated to be a powerful tool to tackle image-to-image translation problems~\cite{isola2017image,zhu2017unpaired,zhu2017toward,choi2018stargan}.
Hallucinating novel views of the subject of a photo can be naturally framed as an image-to-image translation problem. For human subjects, this has been cast to predicting the person's appearance in different poses~\cite{ma2017pose,siarohin2018deformable,zhao2018multi,esser2018variational}.
Fashion and surveillance domains drew most of the attention, with much progress enabled by large real-world datasets providing multiple views of the same subject~\cite{liu2016deepfashion,zheng2015scalable}.\\
For rigid objects instead, this task is usually referred to as \textit{novel 3D view synthesis} and additional assumptions such as object symmetry are taken into account. In point of fact, symmetry is the most common assumption~\cite{zhou2016view,park2017transformation,zhu2018visual} to synthesise disoccluded portions of the object. Starting from a single image, Yang~\etal~\cite{yang2015weakly} showed how a recurrent convolutional network can be trained via curriculum-learning to perform out-of-plane object rotation. In a similar setting Tatarchenko~\etal~\cite{tatarchenko2016multi} predicted both object appearance and depth map from different viewpoints. Successive works~\cite{zhou2016view,park2017transformation} trained a network to learn a symmetry-aware appearance flow, re-casting the remaining synthesis as a task of image completion; \cite{sun2018multi} extends this approach to the case in which $N>1$ input viewpoints are available. However, all these works~\cite{yang2015weakly,tatarchenko2016multi,zhou2016view,park2017transformation,sun2018multi} assume the target view to be known at training time. As this is not usually the case in the real-world, these approaches limit themselves to train solely on synthetic data and exhibit limited generalization in a real-world scenario. The recent work by Zhu~\etal~\cite{zhu2018visual} exploits cycle consistency losses to overcome the need of paired data, thus training on datasets of segmented real-world cars and chairs they gathered for the purpose. Although that work shows more realistic results, it requires pixel-level segmentation for each class of interest. In contrast, we show that already available datasets for object 3D pose estimation~\cite{xiang2014beyond,xiang2016objectnet3d} can be used for this purpose, despite the extremely rough alignment between the annotated model and the image.
\\ \\
\noindent \textbf{3D Shape reconstruction}
Few recent works~\cite{wu2017marrnet, zhu2018visual} have shown that the use of 2.5D sketches can be a viable path to bridge the gap between synthetic and real-world data. In particular, in Zhu \etal~\cite{zhu2018visual} the 2.5D sketch consists of both a silhouette and a depth image rendered from a learnt low-resolution voxel grid by means of a differentiable ray-tracer. While this method is appealing for its geometrical guarantees, it is limited by a number of factors: i) it requires a custom \textit{differentiable} ray-tracing module; 
ii) footprint of voxel-based representations scales with the cube of the resolution despite most of the information lying on the surface~\cite{sinha2017surfnet,palazzi2018end}; iii) errors in the 3D voxel grid naturally propagate to the 2.5D sketch. We also follow this line of work to provide soft 3D priors to the synthesis process. However, in our semi-parametric setting 2.5D sketches are additional inputs which can be rendered from arbitrary viewpoints using standard rendering engines.\\ \\
\noindent \textbf{Nonparametric view synthesis}
In the interactive editing setting, recent works~\cite{kholgade20143d,rematas2017novel} have shown astounding results by keeping the human in the loop and assuming a perfect (even part-level) alignment between the 3D model and the input image. However, as pixels are warped from the input to the target view~\cite{rematas2017novel} it is not feasible to perform shape transfer to a completely different model. Moreover, the time required to synthesise the output is still far from real-time (few seconds). 
On the opposite, our method enables disentangled shape and appearance transfer in real-time, with only a coarse alignment between the input image and the 3D model.\\ \\
In the scenario of image synthesis from semantic layout the recent work of Qi~\etal~\cite{qi2018semi} has shown that non-parametric components (i.e. a memory bank of image segments) can be integrated in a parametric image synthesis pipeline to produce impressive photo-realistic results. Despite our different setting, we similarly rely on image patches to provide hints to the Image Completion Network (\icn); however, our patches are not queried from a database but warped directly from the input view.
\\ \\
After our initial submission, several new related works have been published by the scientific community. We include here a non exhaustive list of them; \cite{nguyen2019hologan,xiao2019identity,sitzmann2019deepvoxels,oechsle2019texture,olszewski2019transformable,chen2019monocular}.
\begin{figure}[t]
    \centering
    \includegraphics[width=0.98\columnwidth]{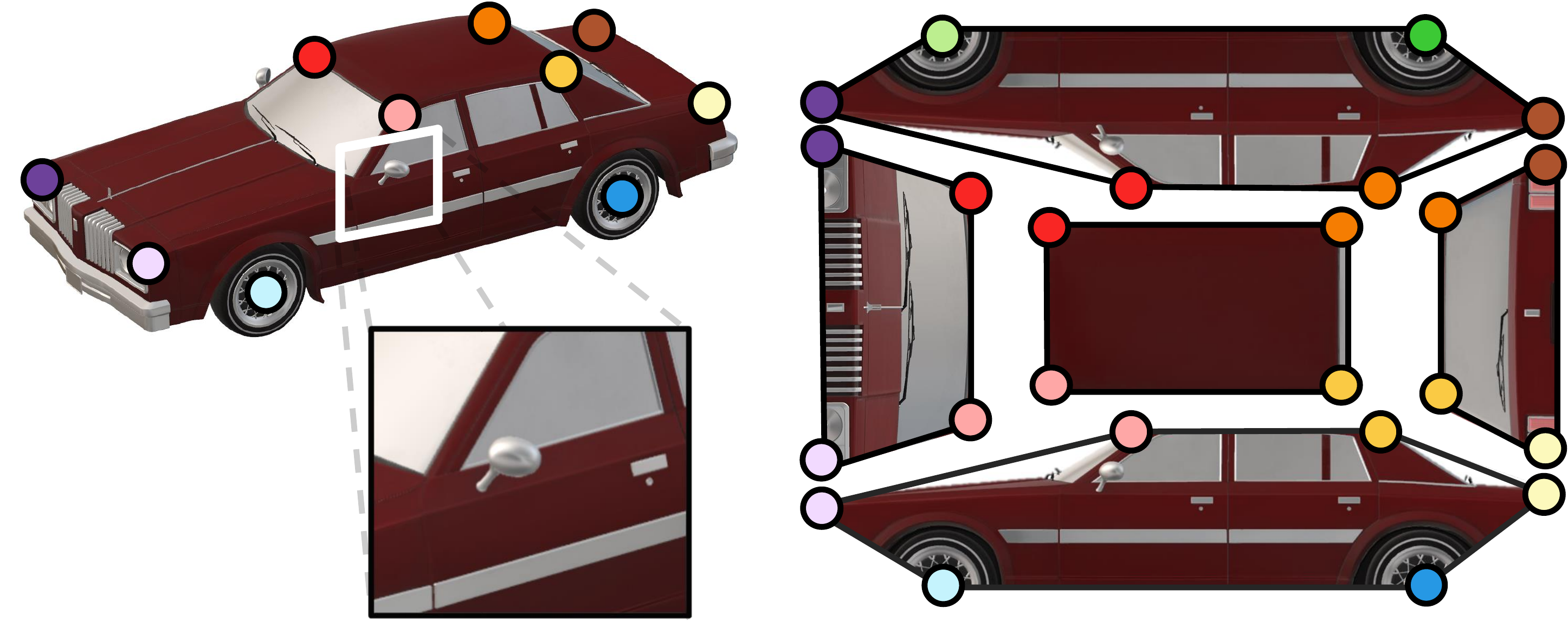}
    \caption{We model a rigid object with a small of piece-wise planar patches, whose vertices are defined by 2D keypoints. We also include a small central crop as appearance prior to carry low-frequency information.}
    \label{fig:car_planes_split}
\end{figure}
\begin{figure*}[ht!]
    \centering
    \includegraphics[width=0.95\textwidth]{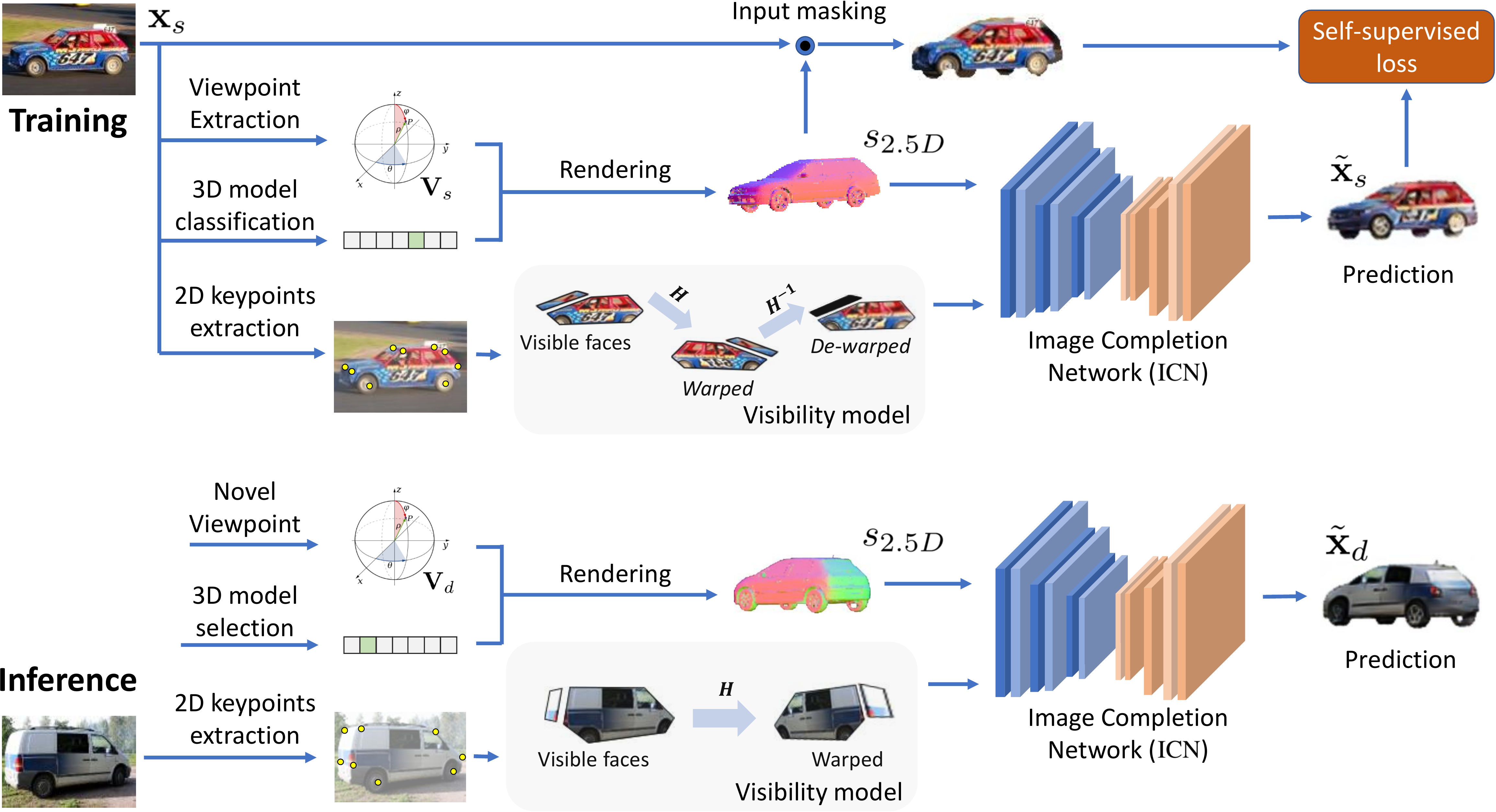}
    \caption{Model architecture overview. Approximately planar patches are extracted from the 2D keypoints locations. The Image Completion Network (\icn) uses the synthetic 2.5D sketches as templates to reconstruct object's appearance from the patches in a self-supervised fashion. During training, input patches are warped forth and back to a randomly sampled viewpoint to enforce resilience against homography issues that are likely to be encountered at test time. During inference, novel views of the input object are synthesised by providing the \icn a novel viewpoint and a (possibly different) rendered 3D model to be used as shape guideline.}
    \label{fig:architecture}
\end{figure*}

\section{Model}
\noindent 
Our model generates novel views of objects in a semi-parametric setting - relying on both geometry and learning. To this end, 2D keypoints and additional 3D information are extracted from a single view of the object. 
Keypoints are used as a proxy to describe 2D geometrical abstractions of the 3D shape (i.e. planar patches), which are transformed to the novel viewpoint. 
Eventually, a convolutional neural network seamlessly fuses this prior information to generate a realistic image from the novel view.\\
More into details, our model takes as input an image depicting a single object $\srcimage \in \imagespace$ viewed from the source viewpoint $\srcviewpoint \in \vpointspace$, its 2D keypoints $\srckpoints$ and its associated 3D CAD model $C \in \meshspace$ having 3D keypoints $\bm{\mathcal{K}}_{3D}$.\\
Training (Fig.~\ref{fig:architecture}, top) is performed in a \textit{self-supervised} fashion maximizing the consistency between the input image and the generated one when projected onto the source viewpoint, with no need for coupled images from the two viewpoints as supervision.
Given $\srcimage$ and $\srckpoints$, planar patches are extracted (Sec.~\ref{subsec:keypoint_decomposition}). The patches are then projected to the target viewpoint $\dstviewpoint$ through an intermediate view (Sec.~\ref{subsec:warping_dewarping}) according to a visibility model  (Sec.~\ref{subsec:visibility}). The 3D model $C$ is also rendered from the target viewpoint $\dstviewpoint$ to get a 2.5D sketch of the object (Sec.~\ref{subsec:leveraging_sketches}). Eventually, the image completion network (\icn) starts from these visual seeds to generate a realistic final image (Sec.~\ref{subsec:appearance_prior},~\ref{subsec:network}).\\
Inference (Fig.~\ref{fig:architecture}, bottom) follows a similar flow. However, in this case only 2D keypoints $\srckpoints$ and source viewpoint $\srcviewpoint$ are needed; the 3D model can be either inferred from the input or arbitrarily selected to perform shape transfer.\\ \\
Without loss of generality, in this work we rely on ground truth data whenever possible, as our focus lies on the overall viewpoint generation pipeline. Off-the-shelf detectors~\cite{he2017mask,pavlakos20176,tulsiani2015viewpoints} can be used to provide these information in an in-the-wild scenario.
\subsection{Keypoint-based decomposition into planar patches}
\label{subsec:keypoint_decomposition}
\noindent We leverage 2D keypoints to approximate the visible shape of the object with a simple polyhedron with a small set of faces, as depicted in Fig.~\ref{fig:car_planes_split}.
Since keypoints mark characteristic locations in the object shape (e.g. corners), a face defined from at least three of those could carry a perceptual / semantic meaning (e.g. the roof of a car).
Exploiting 2D keypoints to find object faces is appealing for a number of reasons. First, this makes straightforward computing the homography matrix between planes in different viewpoints (see Sec.~\ref{subsec:warping_dewarping}). Furthermore, a number of datasets provide object landmark annotations in real-world scenarios~(e.g.~\cite{lpt2013ikea,xiang2014beyond,xiang2016objectnet3d,wang2017veri,wu20183d,huang2018apolloscape}) and solid keypoints detection methods exist~\cite{tulsiani2015viewpoints,pavlakos20176,he2017mask}.\\
Specifically, for each source image $\srcimage \in \imagespace$ an array of 2D keypoints $\srckpoints \in \kpointspace$ is available, being $|\srckpoints|$ the category-specific number of keypoints (we set $|\srckpoints|=12$ for vehicles). From these a set of planar patches are extracted according to the geometry of the object:
\begin{equation}
\srcplset = \{\srcpl^{(0)}, \srcpl^{(1)}, \dots, \srcpl^{(|\mathcal{P}|)}\},\quad \srcpl^{(i)} = \Gamma(\bm{k}_{s}^{(i)})
\end{equation}
where each patch $\srcpl^{(i)}$ is defined as the convex hull $\Gamma$ of the subset of keypoints $\bm{k}_{s}^{(i)} \subseteq \srckpoints$. 
Prior knowledge about the object class can be leveraged to choose the planar patches $\srcplset$: for instance, we choose roof, left, right, front and back sides for vehicles.
\subsection{Warping and dewarping}
\label{subsec:warping_dewarping}
\noindent \textbf{Warping patches} Source patches $\srcplset$ are warped to the destination viewpoint to get a set of warped patches $\dstplset$ that are employed to bootstrap the novel viewpoint synthesis. To this end, we define the destination viewpoint $\dstviewpoint \in \vpointspace$ to be an arbitrary rigid transformation of the camera:
\begin{equation}
    \dstviewpoint = 
    \begin{bmatrix}
    \mathbf{R} & \mathbf{t} \\
    \mathbf{0}^T & 1 
\end{bmatrix}
\end{equation}
Locations of 2D keypoints $\dstkpoints \in \kpointspace$ in the novel viewpoint can be now computed by using the classical pinhole camera model as:
\begin{equation}
    k_d^{(i)} = 
        \begin{bmatrix}
        f & 0 & c_x \\
        0 & f & c_y \\
        0 & 0 & 1
        \end{bmatrix}
\dstviewpoint^{-1} k_{3D}^{(i)}
\end{equation}
where $k_{3D}^{(i)}$ is the $i^{th}$ 3D keypoint in the CAD model and $c_x$, $c_y$ are the principal point coordinates. As the focal $f$ is unknown, we set it to an high value to minimize perspective effects; we choose $f=3000$ as in~\pascal~\cite{xiang2014beyond}. 
A set of homography transformations $\mathbf{H}$ relating planar surfaces in the two views can be estimated from correspondences between $\srckpoints$ and $\dstkpoints$. In this way patches in the destination viewpoint (\textit{\warped} from now on) can be computed as:
\begin{equation}
    \dstplset = \{\dstpl^{(0)}, \dstpl^{(1)}, \dots, \dstpl^{(|\mathbf{P}|)}\},\quad \dstpl^{(i)} = \mathbf{H}^{(i)}( \srcpl^{(i)})
\end{equation}
\noindent \textbf{De-warping patches}
Since real-world datasets do not provide paired views, it is not possible to supervise the destination image $\dstgenimage$; hence, we propose to train the \icn network in a \textit{self-supervised} manner. A straightforward approach would be to reconstruct $\srcimage$ from $\srcplset$. Nonetheless, this would create a distribution shift between the data fed to the network during training and inference stages. In fact, while $\srcplset$ is perfectly aligned with $\srcimage$, $\dstplset$ might be affected by homography failures and interpolation errors among other issues. For example, when the destination area is smaller than the source one many source pixel land onto the same destination pixel, and the inverse warping cannot recover all the information in the original patch. To alleviate this shift, we train the network to reconstruct the image $\srcimage$ from a third set of patches (called \textit{\dewarped} in what follows):
\begin{equation}
    \srcplsetunwarp = \{\srcplunwarp^{(0)}, \srcplunwarp^{(1)}, \dots, \srcplunwarp^{(|\mathbf{P}|)}\},\quad \srcplunwarp^{(i)} = (\mathbf{H}^{(i)})^{-1}(\dstpl^{(i)})
\end{equation}
During training, patches are warped towards a random viewpoint sampled from the training set distribution before being warped back to the source one. In this way the network learns to cope with possible transformation errors and cannot simply short-circuit input patches to the output.
The importance of this \textit{dewarping trick} for a well-behaved network training is highlighted in Sec.~\ref{subsec:visual_results}.
\subsection{Visibility model}
\label{subsec:visibility}
Whenever a 3D object is projected into a 2D image, self-occlusions almost inevitably arise. Consequently, not all planar patches into the set $\srcplset$ are effectively visible. Were they to be warped regardless of their visibility, following parts of our architecture would require to discern which of them to keep or discard. Furthermore, when warping between $\srcplset$ and $\dstplset$ the visibility of some of those planes may vary.
To take these dynamics into account, we first render the object 3D model from the camera viewpoint to obtain the 3D planes corresponding to the detected 2D patches. 
The z-buffer computed through ray-casting is then exploited to filter the patches which are not visible from the source viewpoint.
These are \enquote{dropped}, in the sense that they are zeroed before feeding them to the \icn. As during training the intermediate viewpoint $\dstviewpoint$ is sampled randomly, the warping-dewarping phase results in a random dropout at patch-level, where the chance of drop is inversely proportional to the frequency of visibility of the patch. This forces the network to hallucinate missing patches during training, thus improving generalization when source and destination viewpoint differ.
%
%
%
%
\begin{figure*}[t]
    \centering
    \includegraphics[width=\textwidth]{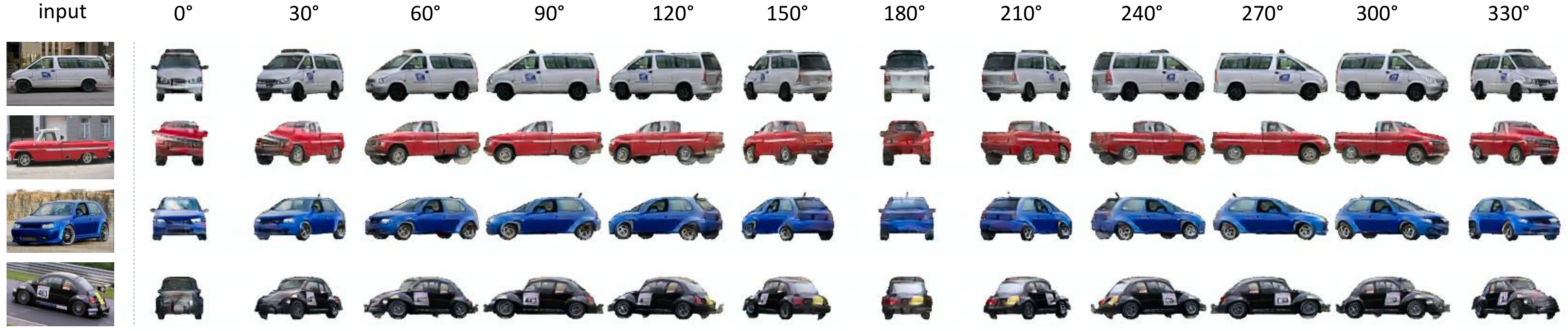}
    \caption{Results of 360° rotation. Our output is consistent for the whole rotation circle. Best viewed zoomed on screen.}
    \label{fig:rotation_results}
\end{figure*}
\subsection{Leveraging 2.5D sketches}
\label{subsec:leveraging_sketches}
While image patches carry rich information about the appearance of the object, they bear few cues about the object shape. In other words, visual aspect and shape are disentangled by design. This is a desirable property enabling multiple applications which require to change one of the two while keeping the other fixed. In this section we propose a method to constrain the synthesised object shape. Let 
\begin{equation}
\cadset = \{C^{(0)}, C^{(1)}, \dots, C^{(|\cadset|)}\}, \quad \cadith \in \meshspace
\end{equation}
be the set of 3D CAD models which approximate the intra-class variation for the current object class, each $\cadith$ being a 3D mesh composed of $f$ faces. The number of CADs $|\cadset|$ needed to cover the intra-class variation reasonably depends on the object category, but it is often relatively low (e.g. $|\cadset|=10$ for the vehicle class in the Pascal3D+ dataset~\cite{xiang2014beyond}). 
Each training example $i$ is thus composed by an image $\mathbf{x}\ith$ and its associated viewpoint $\genericviewpoint\ith$ and CAD index $\alpha \in \{0, 1, \dots, |\cadset|\}$ which can be possibly selected through a classifier. Therefore, a virtual camera can be used to render the CAD $C^{(\alpha)}$ from viewpoint $\genericviewpoint\ith$. In particular, following~\cite{wu2017marrnet}, we render the 2.5D sketch of CAD surface normals:
\begin{equation}
    \sketches ( C^{(\alpha)}, \genericviewpoint\ith) \in \sketchspace
\end{equation}
which provides rich information about the object's 3D shape. During training, this 2.5D sketch is fed to the \icn together with de-warped patches $\srcplsetunwarp$ to reconstruct $\srcimage$.
\subsection{Appearance prior}
\label{subsec:appearance_prior}
\noindent Our method relies on warped patches to transfer the object appearance from a source to a destination viewpoint. Still it might happen that viewpoints $\srcviewpoint$ and $\dstviewpoint$ are so far apart (e.g. front to back) that an object shares no visible faces across the two even with symmetry constraints. To alleviate this issue, we crop from the center of the input image $\srcimage$ a small patch $\srccrop$ with side $10\%$ of the image size and give it as an additional input to the image completion network as a prior knowledge about the rough object appearance in absence of other hints (depicted in Fig.~\ref{fig:car_planes_split}). The network can extract from this crop coarse information about the object visual aspect (e.g. the average color) as a prior to cope with large changes between viewpoints.
%
\subsection{Image Completion Network}
\label{subsec:network}
\noindent The Image Completion Network (\icn) $g(\cdot \given \theta)$ is a fully convolutional network parametrized by $\theta$ trained to reconstruct a realistic image $\srcimage$ from \dewarped~$\srcplsetunwarp$, 2.5D sketch $\sketches$ and appearance prior $\srccrop$:
\begin{equation}
    \srcgenimage = g(\srcplsetunwarp, \sketches(C^{(\alpha)}, \srcviewpoint), \srccrop \given \theta)
\end{equation}
\textbf{Architecture}
Our \icn features an encoder / decoder structure as in \cite{zhu2017unpaired,zhu2018visual}. The encoder is composed of 3 convolutional blocks to reduce the spatial resolution and 3 additional residual blocks applied to the lowest resolution code.
Except for the first one, every convolution has kernel size 4x4 and it's preceded by reflection padding, while ReLU activation and Instance normalization are applied afterwards. The decoder follows the same structure in reverse order.
For the discriminator network we rely on the two-scale PatchGAN classifier~\cite{isola2017image,zhu2017unpaired,zhu2018visual}.\\
%
%
\textbf{Objective}
A number of recent works~\cite{johnson2016perceptual,chen2017photographic,qi2018semi} indicate that loss functions based on high-level features extracted from pretrained networks can lead to much more realistic results compared to naive \textit{per-pixel} losses between the output and ground-truth image.
Given a set of layers $\{\Phi_l\}$ from a network $\Phi$ and a training pair consisting of a real and a generated images $(\srcimage,\srcgenimage)$, we define the perceptual loss function as
\begin{equation}
\mathcal{L}^{VGG}_{\srcimage,\srcgenimage}(\theta) = \sum_l{\lambda_l\| \Phi_l(\srcimage)-\Phi_l(\srcgenimage)\|_1}.
\label{eq:loss}
\end{equation}
We employ each second convolutional layer of each block in VGG-19~\cite{simonyan2014very} as feature extractor $\Phi_l$; $\{\lambda_l\}$ is set as in~\cite{chen2017photographic}.\\ 
As mentioned above, images generated from novel viewpoints $\dstgenimage$ cannot be directly supervised if the dataset does not provide paired views. Nevertheless, we can still enforce the realism of \icn output in an adversarial fashion.
Given a generic image $\genimage$ synthesised by \icn either in the source ($\srcgenimage$) or the destination ($\dstgenimage$) viewpoint, we set up a min-max game as follows:
\begin{equation}
\mathcal{L}^{adv}_{\srcimage,\genimage} = 
\mathbb{E}_{\srcimage}[\log D(\srcimage)] + \mathbb{E}_{\genimage}[\log (1 - D(\genimage))]
\end{equation}
where D is the discriminator network aiming to distinguish between real and synthesised images. Our total loss is defined as:
\begin{equation}
    \mathcal{L} = \mathcal{L}^{VGG}_{\srcimage,\srcgenimage} + \gamma \mathcal{L}^{adv}_{\srcimage,\genimage}
\end{equation}
where $\gamma$ modulates the contribution of the adversarial term.
\begin{figure}[]
    \centering
    \includegraphics[width=\columnwidth]{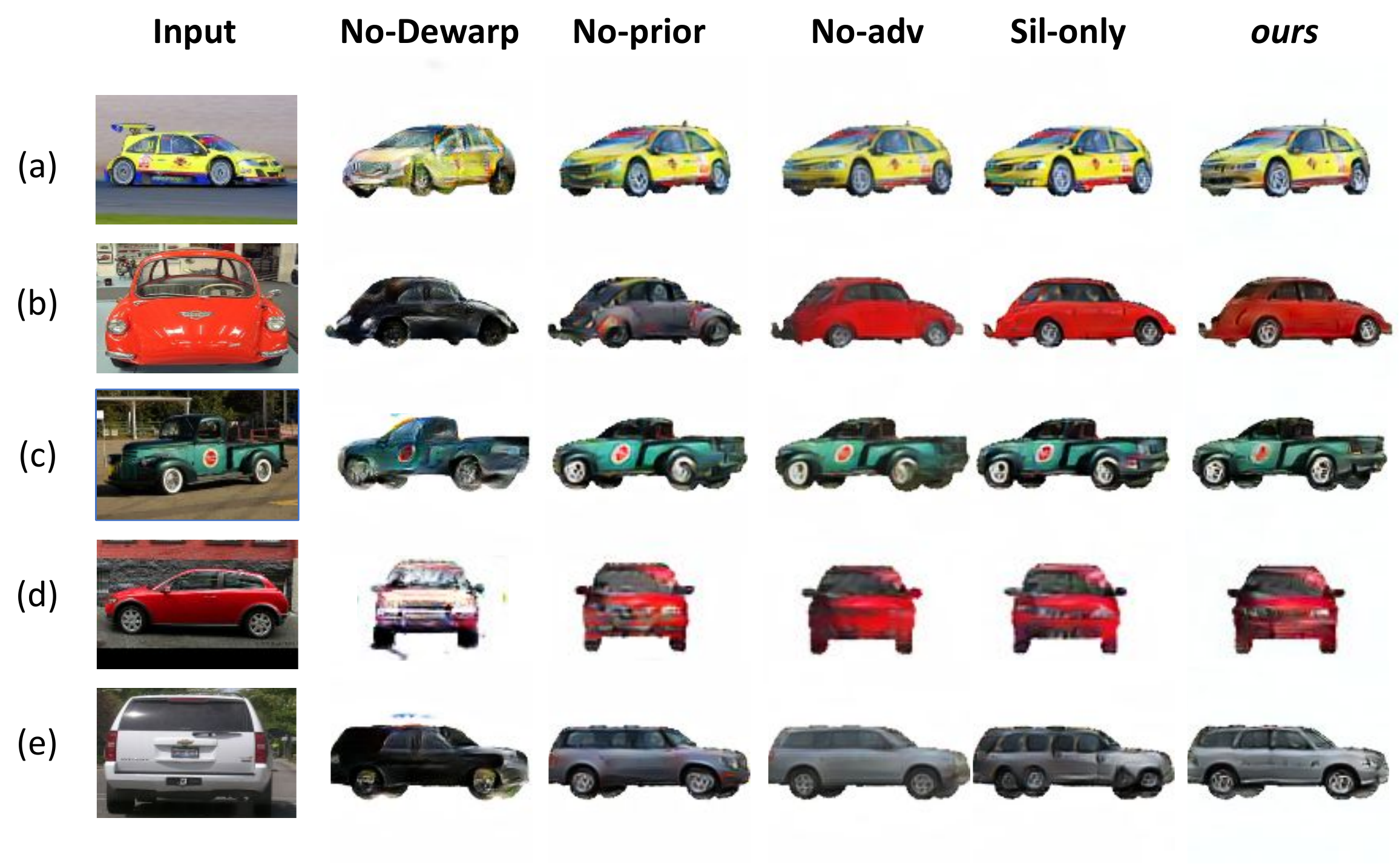}
    \caption{Comparison with ablated versions of the proposed method on \pascal test set. Better viewed zoomed on screen. Please refer to Sec.~\ref{subsec:visual_results} for details.}
    \label{fig:ablation}
\end{figure}
\begin{figure*}[ht]
    \centering
    \includegraphics[width=\textwidth]{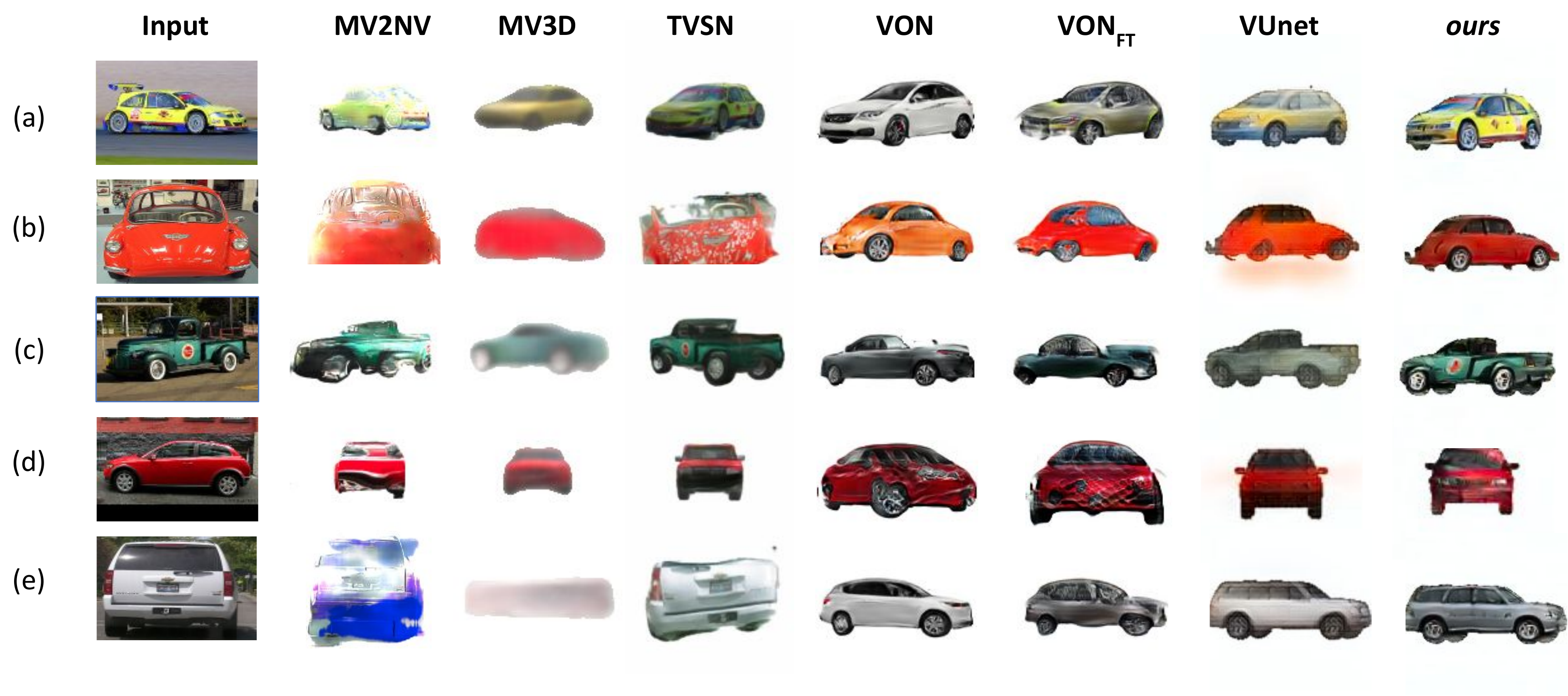}
    \caption{Visual results comparison with competitors on \pascal test set. Better viewed zoomed on screen. Please refer to Sec.~\ref{subsec:visual_results} for details.}
    \label{fig:competitors}
\end{figure*}

\section{Experiments}
\label{sec:experiments}
The experiments we propose concern with the evaluation of the quality of the generated object images across different viewpoints. First, both visual comparison (Sec.~\ref{subsec:visual_results}) and quantitative experiments (Sec.~\ref{subsec:quantitative_results}) against state-of-the-art competitors are reported. Then, we keep the human in the loop by relying on human judgement to measure the output quality via A/B preference tests (Sec.~\ref{subsec:perc_experiments}). Eventually, we extend the evaluation to other classes (Sec.~\ref{subsec:chairs}) and we investigate the contribution of complementary synthetic data for modelling extreme viewpoint changes (Sec.~\ref{subsec:synth}).\\
\begin{figure*}[]
    \centering
    \includegraphics[width=0.9\textwidth]{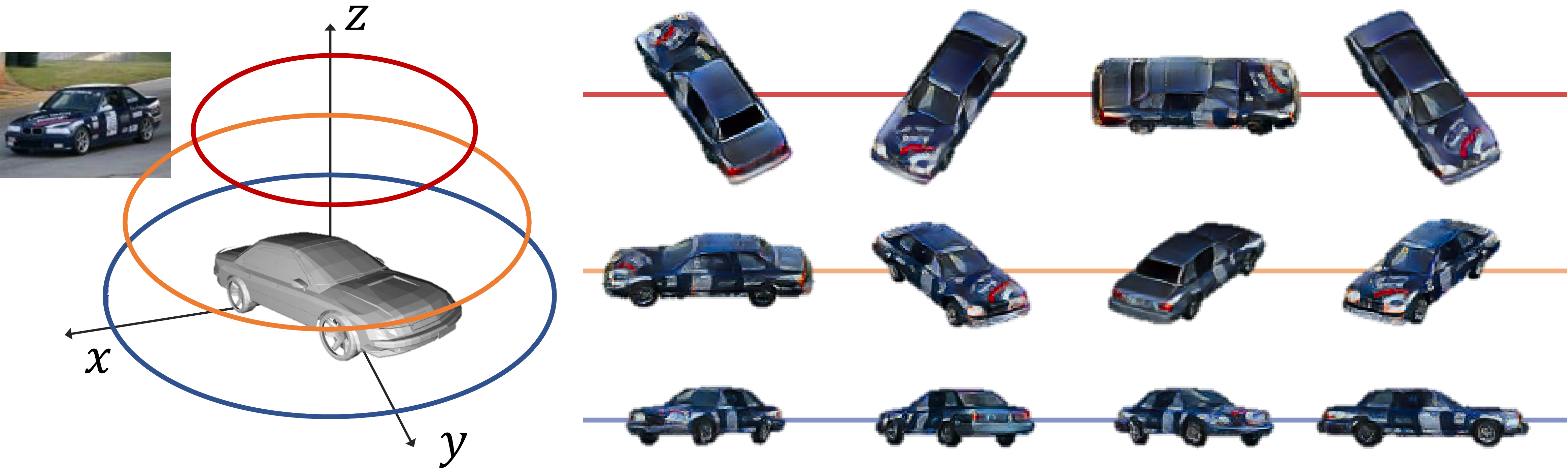}
    \caption{Predictions of our model from different viewpoints. The geometry-aware design of our semi-parametric method allows the model to be resilient to large viewpoint variations, including rotation, elevation and camera distance.}
    \label{fig:cameras}
\end{figure*}
\begin{table*}[]
\begin{center}
\begin{tabular}{l|llllllllllllll}
    \toprule
    & 0° & 30° & 60° & 90° & 120° & 150° & 180° & 210° & 240° & 270° & 300° & 330° & Avg\\
    \midrule
\textbf{ours\textsubscript{real+synth}} & 174.2 & 57.0 &\textbf{ 47.9} & 53.8 & \textbf{52.9} & 59.9 & 168.8 & \textbf{61.2} & \textbf{52.1} & 54.2 & 46.8 & 60.2 & \textbf{74.1}\\
ours\textsubscript{real} & 178.9 & \textbf{54.6} & 51.0 & \textbf{53.2} & 56.8 & \textbf{58.1} & 202.5 & 62.2 & 54.1 & \textbf{49.8} & \textbf{46.3} & \textbf{56.1} & 77.0\\
\MvToNv~\cite{sun2018multi} & 144.5 & 170.9 & 136.5 & 192.3 & 144.7 & 163.1 & 152.2 & 179.1 & 137.5 & 186.0 & 136.0 & 161.3 & 158.7\\
\MvThreeD~\cite{tatarchenko2016multi} & 257.5 & 263.9 & 259.7 & 284.2 & 273.0 & 271.9 & 267.7 & 263.7 & 261.5 & 277.9 & 266.5 & 265.3 & 267.7\\
\tvsn~\cite{park2017transformation} & \textbf{70.8} & 71.3 & 74.1 & 78.9 & 79.0 & 82.2 & \textbf{89.4} & 81.1 & 79.0 & 78.9 & 78.0 & 72.9 & 78.0\\
\vunet~\cite{esser2018variational} & 202.4 & 90.7  & 79.9  & 88.3  & 78.8  & 96.3  & 203.3 & 94.0  & 77.5  & 85.2  & 82.0  & 92.6  & 105.9\\
\von~\cite{zhu2018visual} & 134.1 & 107.2 & 150.0 & 126.5 & 124.4 & 114.4 & 151.7 & 113.8 & 127.2 & 128.9 & 132.0 & 107.3 & 126.4\\
\vonfinetuned~\cite{zhu2018visual} & 165.2 & 100.6 & 137.8 & 125.9 & 137.6 & 108.1 & 190.5 & 155.7 & 134.8 & 123.1 & 117.1 & 102.1 & 133.2\\
\bottomrule
\end{tabular}
\end{center}
\caption{\fids~\cite{heusel2017gans} results for \textit{car}. Each row lists the average distance between real and generated images for each method on the left. Results are reported from 12 evenly spaced azimuthal angles while rotating around the object at fixed elevation and radius. Details in Sec.~\ref{subsec:quantitative_results}.}
\label{tab:fid_by_angle}
\end{table*}
\noindent\textbf{Datasets}
Although large-scale 3D shape repositories providing object geometries such as Princeton Shape Benchmark~\cite{shilane2004princeton} and Shapenet~\cite{chang2015shapenet} exist, they do not come with real-world images aligned.
As we want to work with real-world data, we rely on \pascal~\cite{xiang2014beyond}, an in-the-wild 3D object detection dataset which augments the 12 rigid categories of the PASCAL Visual Object Classes (VOC)~\cite{everingham2010pascal} with 3D annotation -roughly~\cite{sun2018pix3d}- aligned.
In particular, we use the \textit{car} and \textit{chair} subsets, which consist of around 5000 and 1500 images respectively.\\
\noindent \textbf{Competitors}
We evaluate our method against six state-of-the-art works in the task of novel viewpoint synthesis. The first one is \von~\cite{zhu2018visual}, an adversarial learning framework in which object shape, viewpoint and texture are treated as three conditionally independent factors that contribute to the synthesis of the novel viewpoint. Since \von was originally trained on a custom car dataset collected by the authors~\cite{zhu2018visual}, for a fair comparison we implement a second baseline \vonfinetuned by fine-tuning their network on \pascal. For both competitors we provide ground truth voxelized shapes from ~\cite{zhu2018visual} matching the images' CADs, relying on the texture encoder network from~\cite{zhu2018visual} for extracting the textures from \pascal images.
Third, we compare to \vunet~\cite{esser2018variational}, a state-of-the-art framework for conditional image generation based on variational autoencoder~\cite{kingma2013auto}, which shows a good generalization capability across a variety of poses and viewpoints.
In the authors' implementation~\cite{esser2018variational} a U-net~\cite{ronneberger2015unet} architecture is fed with keypoint-based skeletons to perform pose-guided human generation. We re-train their model on \pascal to perform pose-guided object generation. For this process, we feed their shape-encoding network with our 2.5D sketches rendered from the object CAD - which is a more informative signal than the one used in the original implementation (i.e. skeleton or edges) as it also includes the direction of the normal surfaces and fine-grained details from the 3D model.\\
We also compare with three recent \textit{pairwise-trained} models: \MvThreeD~\cite{tatarchenko2016multi}, \tvsn~\cite{park2017transformation} and \MvToNv~\cite{sun2018multi}. Pairwise methods share the need for both source and target pairs during training; thus we cannot re-train or finetune them on \pascal and we rely on pre-trained models released by the authors.\\
To maximize evaluation fairness, in what follows we only sample novel viewpoints rotating around the z-axis at fixed distance and elevation, which is the only setting handled by competitors. Still, our method can handle general roto-translations as well as variation in camera intrinsic.
Fig.~\ref{fig:cameras} and Fig.~\ref{fig:synth_compare} show visual results for large viewpoint changes in both elevation and azimuth; even more extreme transformations are depicted in Fig.~\ref{fig:armchair_backflip}.\\
\noindent \textbf{Implementation details} The 2D bounding box of each example of \pascal\cite{xiang2014beyond} is padded to a squared aspect ratio and resized to 128x128 pixels. We work in LAB space relying on the training procedure from~\cite{xian2018texturegan}. Following \cite{tulsiani2015viewpoints,pavlakos20176} truncated and occluded objects are discarded, resulting in 4081 training and 1042 testing examples respectively. For vehicles, the ours\textsubscript{real+synth} model is trained for 20 epochs with batch size 8. During training images undergo small random rotations, translations and shearing for data augmentation purposes. We use Adam~\cite{kingma2014adam} optimizer with constant learning rate $2 e^{-4}$. Loss balancing term $\gamma$ is set to $8$. The code is developed in PyTorch~\cite{paszke2017pytorch}: we depend on Open3D library~\cite{zhou2018open3d} for 3D data manipulation and rendering. Random search has been employed for hyper-parameter tuning. Without any optimization, inference for a 128x128 image takes $\sim 3 ms$ on a NVIDIA GTX 1080, making it suitable for real-time applications.
\subsection{Visual results}
\label{subsec:visual_results}
\noindent Our model produces high-quality results for a variety of camera viewpoints, preserving fine-grained object appearance, as it can be appreciated in Figures \ref{fig:teaser},~\ref{fig:rotation_results},~\ref{fig:ablation},~\ref{fig:competitors}, \ref{fig:cameras}, \ref{fig:chairs_comparison} and \ref{fig:armchair_backflip}.\\
\noindent \textbf{Competitors} The key differences between the proposed method and competitors can be appreciated in Fig.~\ref{fig:competitors}. From left to right. 
\MvToNv~\cite{sun2018multi} seems to suffer the most the reality gap as well as the lack of multiple views, leading to the worse quality results in our setting. \MvThreeD~\cite{tatarchenko2016multi} predicts at least a reasonable overall shape, although images are blurry to the point that in some cases one can barely recognize the vehicle. Results from \tvsn~\cite{park2017transformation} show the highest variability: while looking generally fine, they are disastrous for less common poses such as (b) and (e). The output from Visual Object Networks~\cite{zhu2018visual} (\von) is generally realistic, but hardly reflects the visual appearance of the input. Furthermore, both \von and \vonfinetuned generator networks do not generalize to poses which are less common in the training set such as the frontal pose in (d). VUnet~\cite{esser2018variational} suffers from blurred results typical from variational autoencoders~\cite{goodfellow2016deep}; also, due to skip connections, input appearance may leak to the output when the two viewpoints are very different (b). 
More generally, the drawbacks of a solely learning-based viewpoint synthesis are evident in (a, c): complex textures cannot be recovered once compressed in a feature vector.\\
\noindent \textbf{Ablation} Ablated versions of our model are shown in Fig.~\ref{fig:ablation}. 
The effect of removing the appearance prior is showcased in \textit{No-prior} column. Without prior information, the \icn fails to infer the object appearance when no planar patch is provided, as shown in (b, e). Removing the adversarial term (\textit{No-adv} column) results in slightly blurred outputs.
We also investigate the aid of the \textit{dewarping trick} presented in Sec.~\ref{subsec:warping_dewarping}. In \textit{No-dewarp} column, the \icn was trained to reconstruct the image from $\srcplset$ instead of $\srcplsetunwarp$ (see Sec.~\ref{subsec:keypoint_decomposition}). As expected, despite the very low reconstruction error at training time (due to the similarity between $\srcimage$ and $\srcplset$), the model fails to generalize to the synthesis of novel viewpoints where the textures $\dstplset$ are the result of an homography transformation. This highlights the importance of the \textit{dewarping trick} for a well-behaved training.
Eventually, \textit{Sil-only} shows the ablated versions in which the input sketch is constituted only by 2D silhouette. Although results do not differ dramatically, it can be appreciated how the network benefits from additional information to resolve ambiguous situations such as self-occlusions (e) and details such as side windows, lights, wheels.\\
\noindent \textbf{Shape transfer} Visual results for shape transfer are showed in Fig.~\ref{fig:teaser}. In this setting, the network is requested to complete the warped faces $\dstplset$ using the 2.5D sketch rendered from a totally different CAD. It can be appreciated how novel viewpoints are still realistic, since the network exploits the 2.5D sketch to complete the warped appearance in a CAD-agnostic manner.
\subsection{Metrics evaluation}
\label{subsec:quantitative_results}
\noindent \textbf{\fid} To quantitatively measure the similarity between generated and real images we rely on \fid (FID), which was shown to consistently correlate with human judgment~\cite{heusel2017gans,lucic2018gans}. We employ activations from the last convolutional layer of an InceptionV3 model pretrained on ImageNet~\cite{deng2009imagenet} as features.
Assuming a multidimensional Gaussian distribution for these features, we compute the FID as follow:
\begin{equation}
\mathrm{FID} =\|\bm{m}-\bm{m}_w\|_2^2+  \mathrm{Tr} \bigl(\bm{C}+\bm{C}_w-2\bigl(
\bm{C}\bm{C}_w\bigr)^{1/2}\bigr)
\end{equation}
Where $\bm{m}$, $\bm{C}$ are the mean and covariance of the features extracted from \pascal data, while $\bm{m}_w$ and $\bm{C}_w$ are the corresponding statistics extracted from the generated images.\\
To enable the comparison with other works, we sample novel viewpoints while rotating around the object at fixed distance and elevation: results are reported in Table~\ref{tab:fid_by_angle}, binned in 12 equidistant azimuthal angles. \fid rewards the realism we can get with our semi-parametric approach; our method outperforms all competitors for the vast majority of viewpoints. In particular, our method preserves both low (i.e. shape) and high-frequency (i.e. texture) image statistics which are both contribute to the overall FID score.
%
\subsection{Perceptual experiments}
\label{subsec:perc_experiments}
\begin{figure}[b]
    \centering
    \includegraphics[width=0.99\columnwidth]{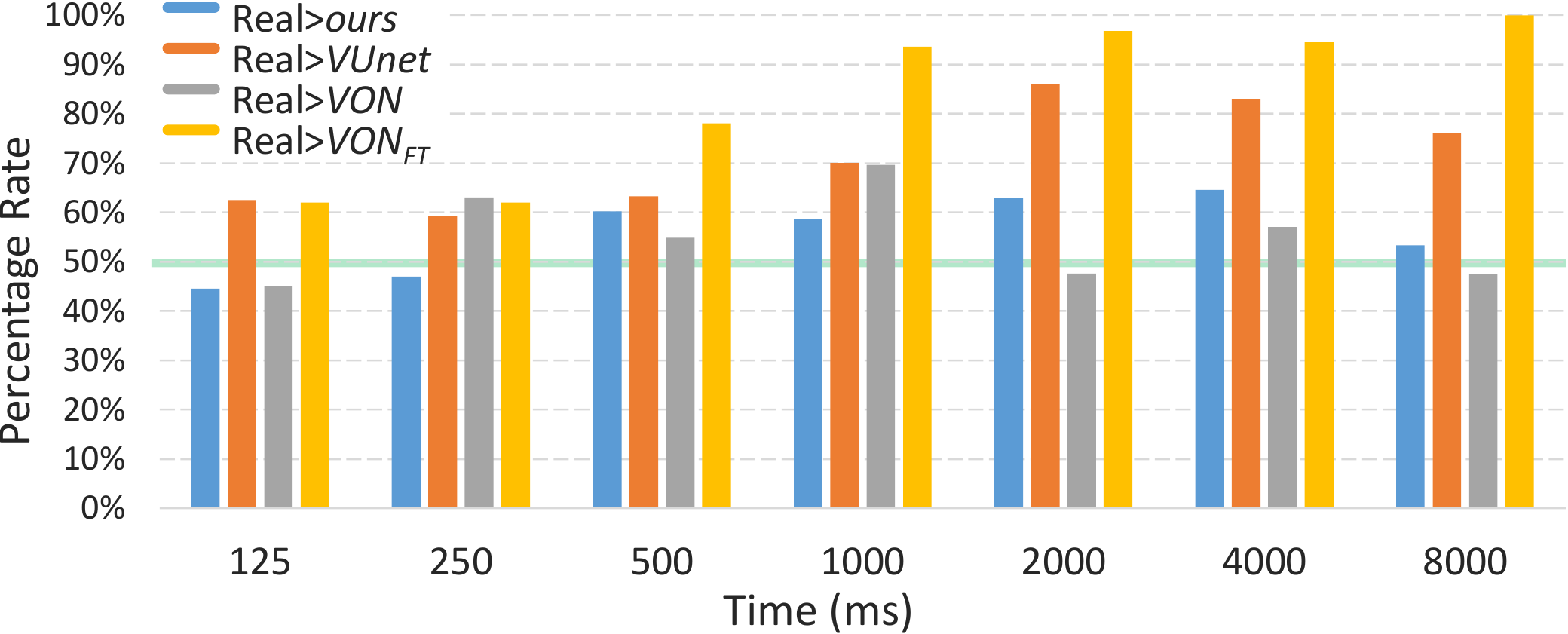}
    \caption{Results of time-limited \textit{A}/\textit{B} preference test against real images. Both \von and our method are resilient to human judgement over time. Green line denotes random chance. Please refer to Sec.\ref{subsec:perc_experiments} for details.}
    \label{fig:percept_exp_graph}
\end{figure}
\begin{table}[]
\begin{center}
\begin{tabular}{@{}l@{}ccc@{}}
    \toprule
    & Car (plain) & Car (textured) & Avg\\
    \midrule
    ours > \vunet~\cite{esser2018variational}           &  76.0\%  &  85.0\%  &  78.0\% \\
    ours > \von~\cite{zhu2018visual}                    &  88.0\%  &  98.0\%  &  91.0\% \\
    ours > \vonfinetuned~\cite{zhu2018visual}           &  96.0\%  &  99.0\%  & 97.0\% \\
    \bottomrule
\end{tabular}
\end{center}
\caption{Blind randomized A/B test results. Each row lists the percentage of workers who preferred the novel viewpoint generated with our method with respect to each baseline (chance is at 50\%).}
\label{tab:ab_tests}
\end{table}
%
\begin{figure*}[]
    \centering
    \includegraphics[width=0.92\textwidth]{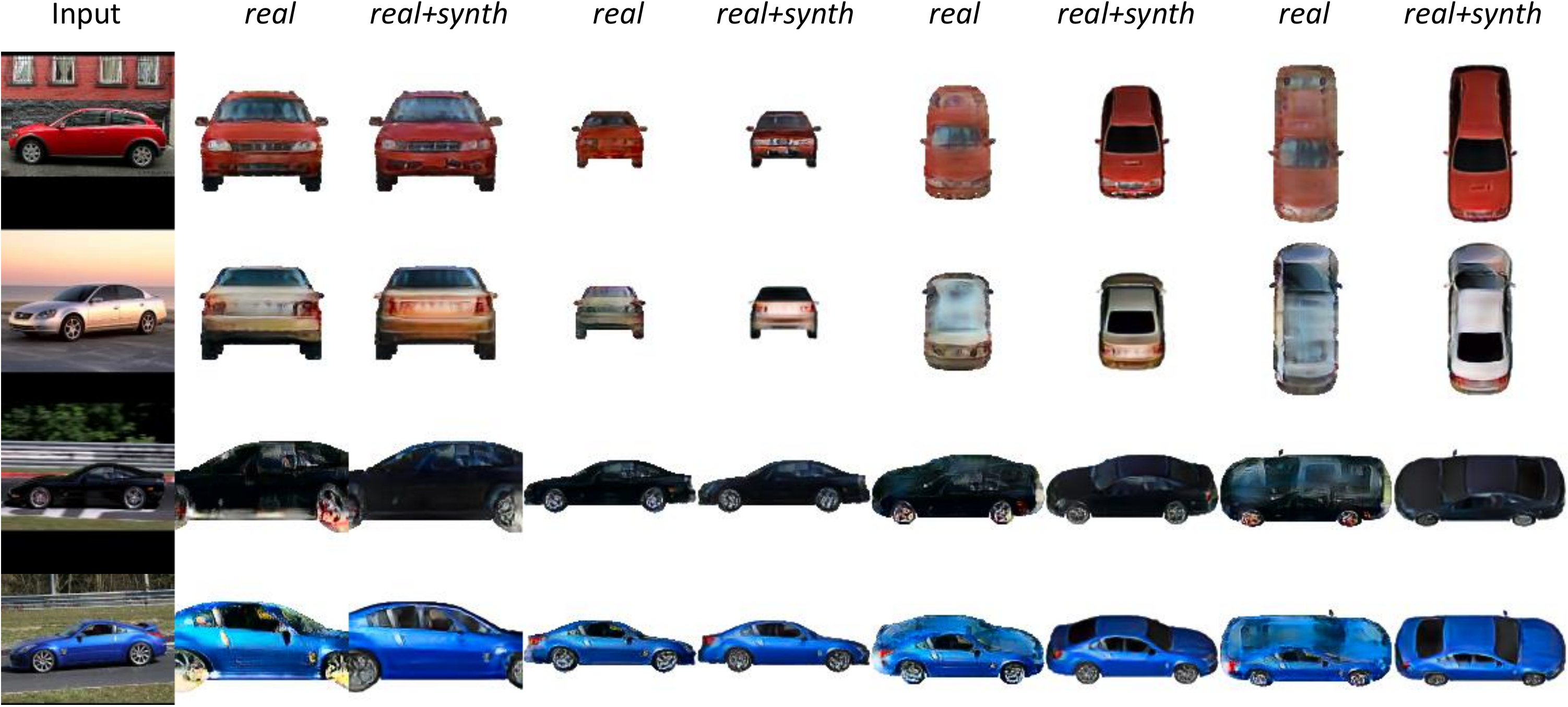}
    \caption{Visual comparison showing the effect of adding synthetic data to \pascal training set. The \icn network trained on the mixture of the two domains performs significantly better under extreme viewpoint transformations. Please see supplementary material for more examples.}
    \label{fig:synth_compare}
\end{figure*}
\noindent To assess the quality of our results also from a perceptual point of view, randomized A/B preference tests were performed by 43 human workers, following the experimental protocol of previous works~\cite{chen2017photographic,qi2018semi,zhu2018visual}. Due to time constraint, for this phase we select only the three most recent competitors, namely \von~\cite{zhu2018visual}, \vonfinetuned~\cite{zhu2018visual} and \vunet~\cite{esser2018variational}. As we want to evaluate both the realism and the appearance coherence of our method, we perform two different tests.\\
\textbf{View transfer coherence} In the first setting, the subject is presented with three images: while the first one comes from Pascal3D+ test set, \textit{A} and \textit{B} depict a novel viewpoint of the object generated with two different methods. The human worker is then asked whether rotating the input object would better lead to \textit{A} or \textit{B}.
Results reported in Tab.~\ref{tab:ab_tests} indicate that our method is largely preferred to competitors, likely because of the built-in realism that comes from warping the original image. As a further analysis we split by manual annotation \pascal images into \plain and \textured sets, the latter set containing vehicles which feature characteristic textures. Table~\ref{tab:ab_tests} highlights that 
workers expressed almost unanimous preference for our method on the \textured set. The fact that human attention was caught by these appearance details highlights the importance of preserving fine-grained details in the synthesized output.\\
\textbf{Output realism} The second experiment consists of a two-alternative forced choice aimed at evaluating the relative realism of each method. Here the subject is presented with only two images for a determined amount of time. The worker is then asked which of the two appeared more realistic and the experiment is repeated by varying the amount of time. Results depicted in Fig.~\ref{fig:percept_exp_graph} follow two trends. On the one hand, workers clearly discern \vunet and \vonfinetuned images from real ones as more time is available. \vunet is hurt by excessive blur and visual artifacts; \vonfinetuned suffers from a severe loss of realism w.r.t. the original \von method, which may be related to the great variety of viewpoints in the \pascal dataset compared with the one used in Zhu~\etal~\cite{zhu2018visual}. On the other hand, both \von and our method produce realistic images workers struggle to distinguish from the real ones even in 8000ms.
%

\subsection{Evaluation on other classes}
\label{subsec:chairs}
\begin{figure}[]
    \centering
    \includegraphics[width=0.9\columnwidth]{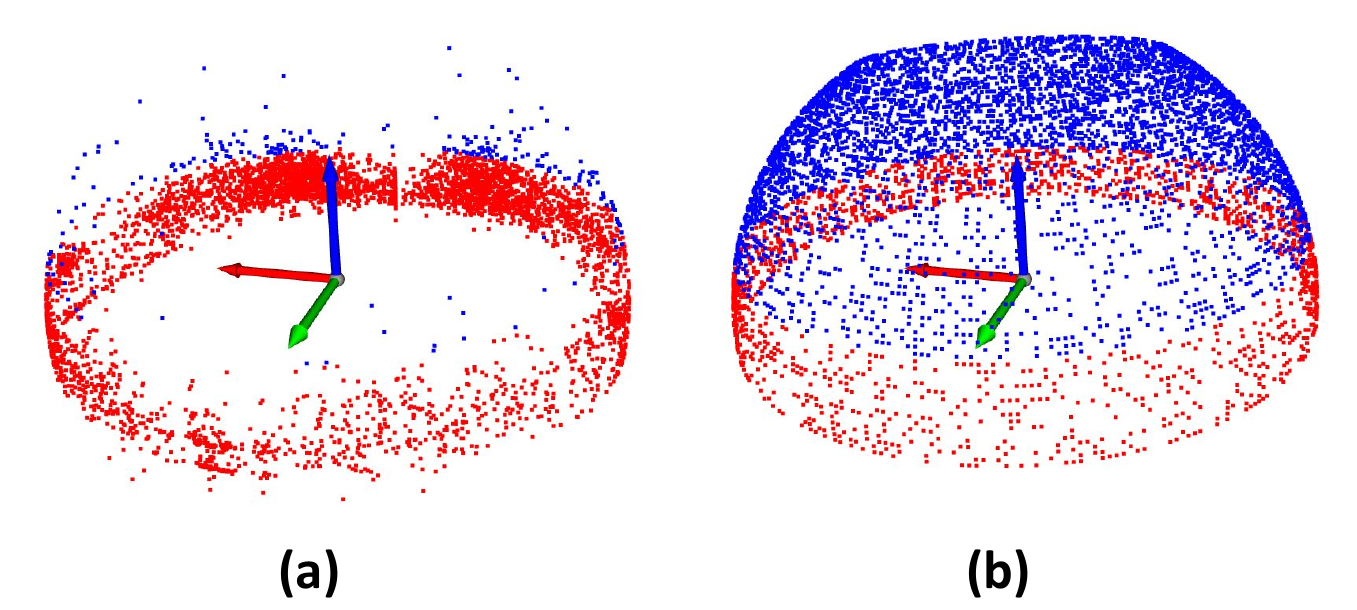}
    \caption{Viewpoints' distributions for real (a) and synthetic data (b) of the car class in \pascal. Radii have been normalised to unit length for clearness. In red viewpoints with elevation lesser or equal than $\frac{\pi}{8}$ rad.}
    \label{fig:viewpoint_dist}
\end{figure}
\begin{figure*}[]
    \centering
    \includegraphics[width=0.82\textwidth]{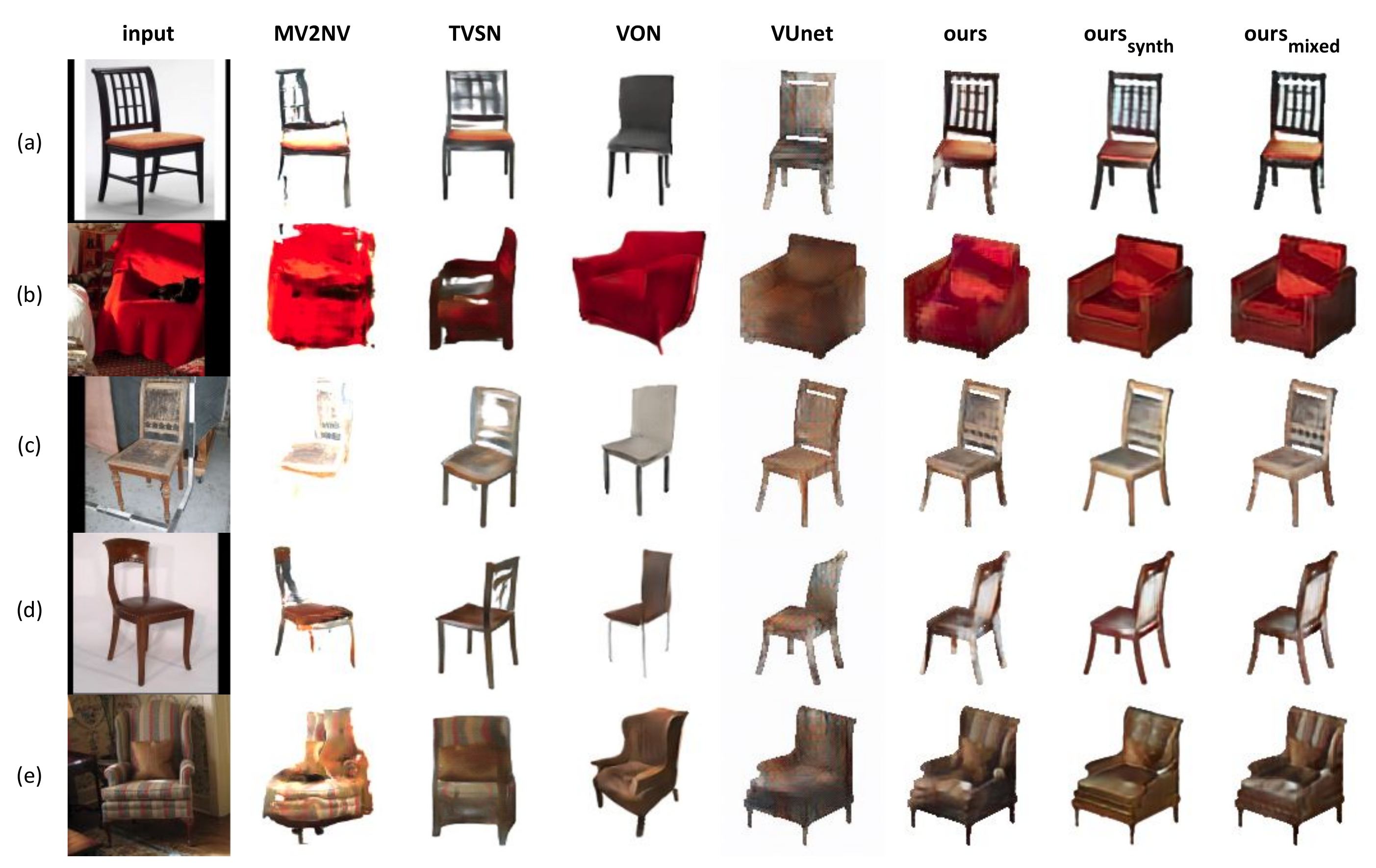}
    \caption{Visual results comparison with competitors for \textit{chair} class on \pascal test set. Better viewed zoomed on screen.}
    \label{fig:chairs_comparison}
\end{figure*}
\begin{figure*}[]
    \centering
    \includegraphics[width=0.82\textwidth]{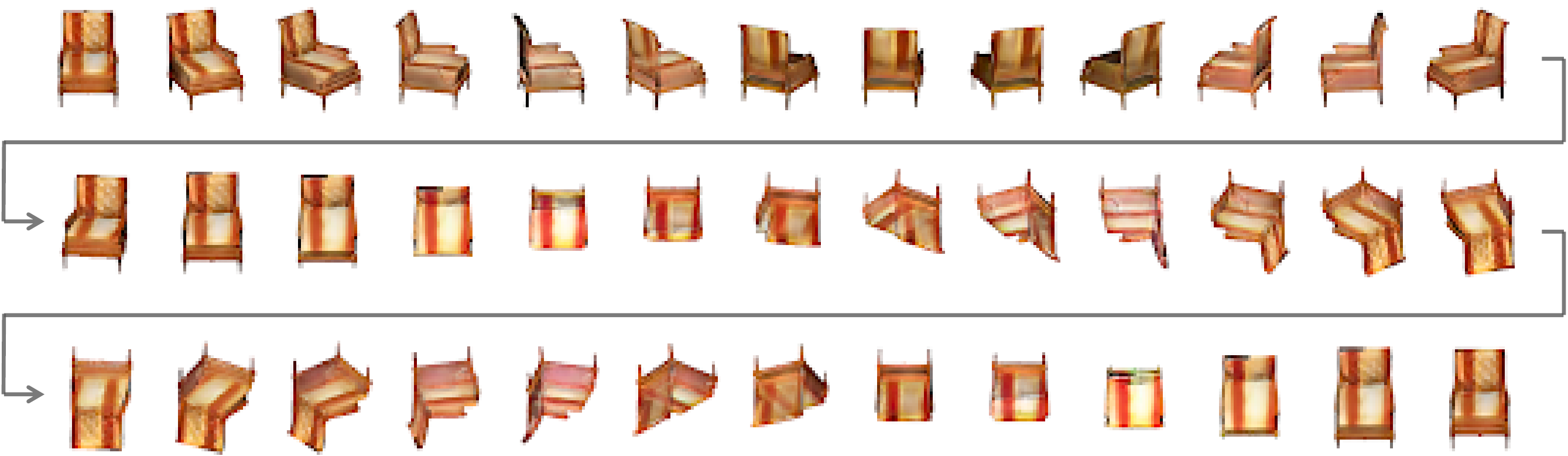}
    \caption{Geometric guidance built in our semi-parametric model allows to perform extreme object transformations which are currently unfeasible for any fully-parametric method. In this case we are able to rotate the armchair upside-down, although this configuration never appears in the training set. Best viewed zoomed on screen.}
    \label{fig:armchair_backflip}
\end{figure*}
\begin{table*}[ht]
\begin{center}
\begin{tabular}{l|llllllllllllll}
    \toprule
    & 0° & 30° & 60° & 90° & 120° & 150° & 180° & 210° & 240° & 270° & 300° & 330° & Avg\\
    \midrule
\textbf{ours\textsubscript{real+synth}} & \textbf{83.9} & \textbf{57.9} & \textbf{66.5} & 120.1 & \textbf{82.5} & \textbf{76.1} & 102.3 & \textbf{77.2} & \textbf{83.4} & 117.0 & \textbf{65.2} & \textbf{60.9} & \textbf{82.8}\\
ours\textsubscript{real} & 92.6 & 62.4 & 68.0 & 122.5 & 84.8 & 84.7 & 117.3 & 82.3 & 85.0 & 127.5 & 70.1 & 64.0 & 88.4\\
\tvsn~\cite{park2017transformation} & 84.7 & 86.6 & 90.8 & \textbf{95.1} & 94.0 & 97.2 & \textbf{93.9} & 94.6 & 95.0 & \textbf{93.0} & 87.8 & 82.7 & 91.6\\
\von~\cite{zhu2018visual} & 125.5 & 107.7 & 100.1 & 121.1 & 122.6 & 136.8 & 203.9 & 141.4 & 123.8 & 114.2 & 102.1 & 96.1 & 124.6\\
\vunet~\cite{esser2018variational} & 118.9 & 97.1 & 123.1 & 171.0 & 160.5 & 137.9 & 151.2 & 141.9 & 155.8 & 154.9 & 120.1 & 95.8 & 135.7\\
\MvToNv~\cite{sun2018multi} & 180.3 & 168.0 & 178.3 & 187.8 & 184.1 & 177.8 & 184.7 & 184.3 & 200.2 & 191.0 & 184.7 & 166.4 & 182.3\\
\bottomrule
\end{tabular}
\end{center}
\caption{\fids~\cite{heusel2017gans} results for \textit{chair} class. Each row lists the average distance between real and generated images for each method on the left. Results are reported from 12 evenly spaced azimuthal angles while rotating around the object at fixed elevation and radius.}
\label{tab:fid_by_angle_chair}
\end{table*}
Similarly to \cite{sun2018multi,park2017transformation, zhu2018visual}, we test our method also on the \textit{chair} subset from \pascal dataset, consisting of 1195 images annotated with 10 different CAD models to assess the generalisation capability of our method.
As for vehicles, also for chairs we define a set of planes from 3D annotated keypoints to approximate the surface of the objects, namely: left, right, seat and back. Even though chairs often feature holes and slits leaking some of the background, we treat them as filled planes. This is due to having access to the foreground segmentation mask through the rendered CAD model, which can be used to mask out background areas. However, the very coarse alignment of the chairs CAD models in \pascal leads to an additional difficulty when training the \icn.\\
Table~\ref{tab:fid_by_angle_chair} reports results against competitors in terms of \fid score: \MvThreeD~\cite{tatarchenko2016multi} is omitted since it only releases pre-trained model for cars. Our method outperforms all other competitors also for this class of objects; visual examples are reported in Fig.~\ref{fig:chairs_comparison}. While MV2NV~\cite{sun2018multi} and \tvsn~\cite{park2017transformation} often struggle to generate the object from the correct viewpoint, VON~\cite{zhu2018visual} fails to transfer visual details in the final output. Although the generated image looks realistic, it doesn't resemble the input one. Contrarily, our method successfully generate realistic views of the input object, even under severe viewpoint transformations~\ref{fig:armchair_backflip}. 
Still, when background leaks in the input mask due to the CAD misalignment (first and fourth row) the final output quality decreases.
\subsection{On the use of synthetic data}
\label{subsec:synth}
Even though the semi-parametric nature of our proposed method copes with a variety of viewpoint, it still rely on data to learn how to stitch together the warped patches. Therefore, for dramatic changes of viewpoint that are completely uncovered in the dataset, performances may drop significantly. As shown in Fig.~\ref{fig:viewpoint_dist}~(a), \pascal viewpoints' distribution for the car set is profoundly skewed towards low elevation values and is polarized with regard to the azimuth to frontal and lateral views, reflecting how the images were acquired.
As it is of great interest to produce realistic images from more varied viewpoints (e.g. bird's eye view), we include synthetic data in the training set for balancing the viewpoints' distribution.
To this end, we sample 59 models from the car synset in the ShapeNet~\cite{chang2015shapenet} dataset and we annotate them with 3D keypoints to define the planes of interest. We then render 6950 images sampling viewpoints uniformly in a 3D semi-spheres around the origin. 
As shown in Fig.~\ref{fig:viewpoint_dist}~(b) this viewpoints' distribution is much more uniformly distributed in terms of azimuth and elevation.
We name ours\textsubscript{real+synth} the \icn network trained on a mixture of synthetic and real data. We also experimented pre-training our model on synthetic data and fine-tuning on the real ones, although in our experience that policy led to worse results.\\
Visual comparison between ours\textsubscript{real} and ours\textsubscript{real+synth} is shown in Fig.~\ref{fig:synth_compare}. It can be seen how training only on \pascal entails artefacts for out-of-distribution viewpoints (e.g. bird's eye view). Conversely, combining the two domains the network learns from synthetic data a prior about the overall structure and color.\\
We perform an analogous augmentation of the chair class. In this case we included 1858 freshly rendered synthetic images from 73 annotated models. As the rendered images are perfectly aligned with the foreground mask, this also contributes to reduce the planes' misalignment introduced by images from \pascal.
%
%
\begin{figure}[]
    \centering
    \includegraphics[width=\columnwidth]{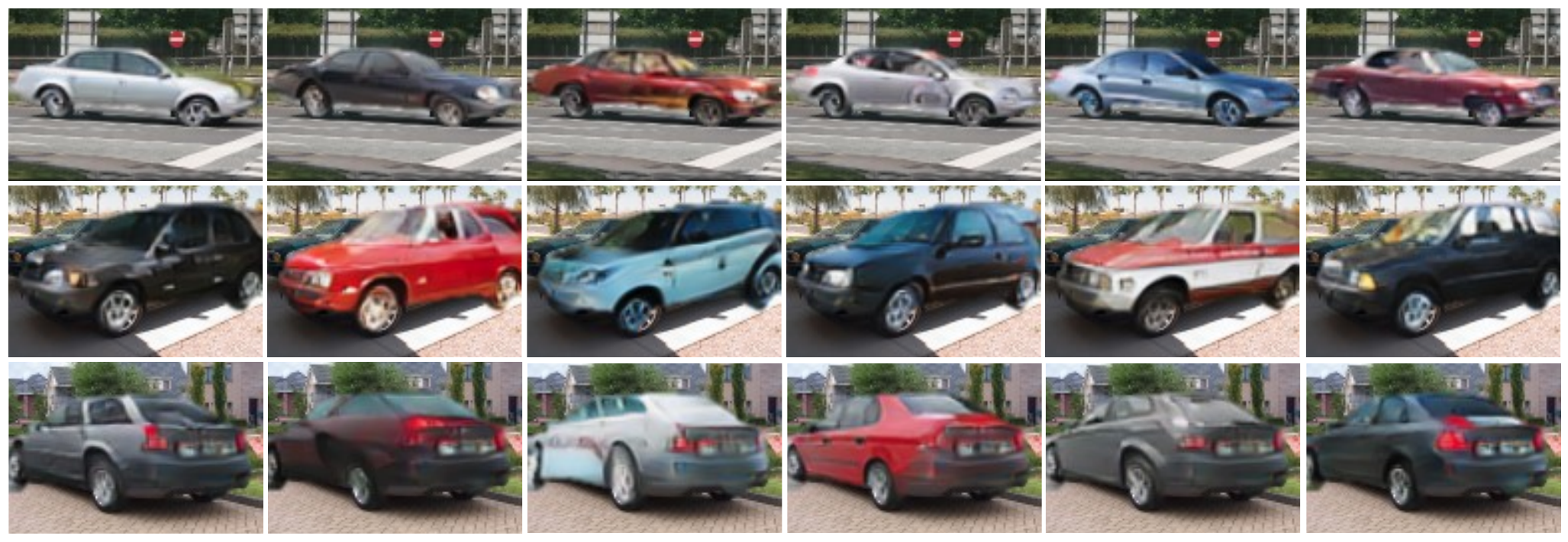}
    \caption{Artificial data created stitching generated vehicles onto \pascal\cite{xiang2014beyond} backgrounds.}
    \label{fig:pascal_aug}
\end{figure}
%

\section{Conclusions}
In this work we introduced a novel formulation of the problem of object novel viewpoint synthesis in a semi-parametric setting.
Notably, our model is designed to be trainable on existing datasets for 3D object detection in a self-supervised manner, without the need for paired source/target viewpoint images - although it can be complemented with synthetic data.
Non-parametric visual hints act as prior information to guide a deep parametric model for generating realistic images, disentangling by design appearance and shape. This enables truly continuous manipulation of the viewpoint and shape transfer to different 3D models.
As completing the image is much easier than generating it from scratch, we can train our \icn on just few thousands images from the \pascal dataset and still be able to generalize to unseen viewpoints.\\
Although vehicles were the main focus of our work, we show that our framework is generic enough to handle rigid objects of completely different geometric structure such as chairs.
Perceptual experiments results as well as image-quality metrics reward our method for its realism and the visual consistency of the synthesised object across arbitrary points of view.

\section*{Acknowledgements}
This research was supported by MIUR PRIN project \textit{"PREVUE: PRediction of activities and Events by Vision in an Urban Environment"}, grant ID E94I19000650001.

%
%
%
%

\ifCLASSOPTIONcaptionsoff
  \newpage
\fi



{\small
\bibliographystyle{ieee}
\bibliography{egbib}

\begin{thebibliography}{10}\itemsep=-1pt

\bibitem{arjovsky2017wasserstein}
M.~Arjovsky, S.~Chintala, and L.~Bottou.
\newblock Wasserstein generative adversarial networks.
\newblock In {\em International Conference on Machine Learning}, pages
  214--223, 2017.

\bibitem{opencv_library}
G.~Bradski.
\newblock {The OpenCV Library}.
\newblock {\em Dr. Dobb's Journal of Software Tools}, 2000.

\bibitem{bundesen1975visual}
C.~Bundesen and A.~Larsen.
\newblock Visual transformation of size.
\newblock {\em Journal of Experimental Psychology: Human Perception and
  Performance}, 1(3):214, 1975.

\bibitem{chang2015shapenet}
A.~X. Chang, T.~Funkhouser, L.~Guibas, P.~Hanrahan, Q.~Huang, Z.~Li,
  S.~Savarese, M.~Savva, S.~Song, H.~Su, et~al.
\newblock Shapenet: An information-rich 3d model repository.
\newblock {\em arXiv preprint arXiv:1512.03012}, 2015.

\bibitem{chaurasia2013depth}
G.~Chaurasia, S.~Duchene, O.~Sorkine-Hornung, and G.~Drettakis.
\newblock Depth synthesis and local warps for plausible image-based navigation.
\newblock {\em ACM Transactions on Graphics (TOG)}, 32(3):30, 2013.

\bibitem{chen2017photographic}
Q.~Chen and V.~Koltun.
\newblock Photographic image synthesis with cascaded refinement networks.
\newblock In {\em IEEE International Conference on Computer Vision (ICCV)},
  volume~1, page~3, 2017.

\bibitem{chen2009sketch2photo}
T.~Chen, M.-M. Cheng, P.~Tan, A.~Shamir, and S.-M. Hu.
\newblock Sketch2photo: Internet image montage.
\newblock In {\em ACM transactions on graphics (TOG)}, volume~28, page 124.
  ACM, 2009.

\bibitem{chen2019monocular}
X.~Chen, J.~Song, and O.~Hilliges.
\newblock Monocular neural image based rendering with continuous view control.
\newblock In {\em Proceedings of the IEEE International Conference on Computer
  Vision}, 2019.

\bibitem{choi2018stargan}
Y.~Choi, M.~Choi, M.~Kim, J.-W. Ha, S.~Kim, and J.~Choo.
\newblock Stargan: Unified generative adversarial networks for multi-domain
  image-to-image translation.
\newblock In {\em Proceedings of the IEEE Conference on Computer Vision and
  Pattern Recognition}, pages 8789--8797, 2018.

\bibitem{deng2009imagenet}
J.~Deng, W.~Dong, R.~Socher, L.-J. Li, K.~Li, and L.~Fei-Fei.
\newblock Imagenet: A large-scale hierarchical image database.
\newblock 2009.

\bibitem{esser2018variational}
P.~Esser, E.~Sutter, and B.~Ommer.
\newblock A variational u-net for conditional appearance and shape generation.
\newblock In {\em Proceedings of the IEEE Conference on Computer Vision and
  Pattern Recognition}, pages 8857--8866, 2018.

\bibitem{everingham2010pascal}
M.~Everingham, L.~Van~Gool, C.~K. Williams, J.~Winn, and A.~Zisserman.
\newblock The pascal visual object classes (voc) challenge.
\newblock {\em International journal of computer vision}, 88(2):303--338, 2010.

\bibitem{feng2018towards}
D.~Feng, L.~Rosenbaum, and K.~Dietmayer.
\newblock Towards safe autonomous driving: Capture uncertainty in the deep
  neural network for lidar 3d vehicle detection.
\newblock In {\em 2018 21st International Conference on Intelligent
  Transportation Systems (ITSC)}, pages 3266--3273. IEEE, 2018.

\bibitem{gardner2011frames}
H.~Gardner.
\newblock {\em Frames of mind: The theory of multiple intelligences}.
\newblock Hachette UK, 2011.

\bibitem{goodfellow2016deep}
I.~Goodfellow, Y.~Bengio, and A.~Courville.
\newblock {\em Deep learning}.
\newblock MIT press, 2016.

\bibitem{goodfellow2014generative}
I.~Goodfellow, J.~Pouget-Abadie, M.~Mirza, B.~Xu, D.~Warde-Farley, S.~Ozair,
  A.~Courville, and Y.~Bengio.
\newblock Generative adversarial nets.
\newblock In {\em Advances in neural information processing systems}, pages
  2672--2680, 2014.

\bibitem{hays2007scene}
J.~Hays and A.~A. Efros.
\newblock Scene completion using millions of photographs.
\newblock {\em ACM Transactions on Graphics (TOG)}, 26(3):4, 2007.

\bibitem{he2017mask}
K.~He, G.~Gkioxari, P.~Doll{\'a}r, and R.~Girshick.
\newblock Mask r-cnn.
\newblock In {\em Proceedings of the IEEE international conference on computer
  vision}, pages 2961--2969, 2017.

\bibitem{hedman2016scalable}
P.~Hedman, T.~Ritschel, G.~Drettakis, and G.~Brostow.
\newblock Scalable inside-out image-based rendering.
\newblock {\em ACM Transactions on Graphics (TOG)}, 35(6):231, 2016.

\bibitem{heusel2017gans}
M.~Heusel, H.~Ramsauer, T.~Unterthiner, B.~Nessler, and S.~Hochreiter.
\newblock Gans trained by a two time-scale update rule converge to a local nash
  equilibrium.
\newblock In {\em Advances in Neural Information Processing Systems}, pages
  6626--6637, 2017.

\bibitem{huang2018apolloscape}
X.~Huang, X.~Cheng, Q.~Geng, B.~Cao, D.~Zhou, P.~Wang, Y.~Lin, and R.~Yang.
\newblock The apolloscape dataset for autonomous driving.
\newblock In {\em Proceedings of the IEEE Conference on CVPR Workshops}, pages
  954--960, 2018.

\bibitem{isola2013scene}
P.~Isola and C.~Liu.
\newblock Scene collaging: Analysis and synthesis of natural images with
  semantic layers.
\newblock In {\em Proceedings of the IEEE International Conference on Computer
  Vision}, pages 3048--3055, 2013.

\bibitem{isola2017image}
P.~Isola, J.-Y. Zhu, T.~Zhou, and A.~A. Efros.
\newblock Image-to-image translation with conditional adversarial networks.
\newblock In {\em 2017 IEEE Conference on Computer Vision and Pattern
  Recognition (CVPR)}, pages 5967--5976. IEEE, 2017.

\bibitem{johnson2016perceptual}
J.~Johnson, A.~Alahi, and L.~Fei-Fei.
\newblock Perceptual losses for real-time style transfer and super-resolution.
\newblock In {\em European conference on computer vision}, pages 694--711.
  Springer, 2016.

\bibitem{karras2017progressive}
T.~Karras, T.~Aila, S.~Laine, and J.~Lehtinen.
\newblock Progressive growing of gans for improved quality, stability, and
  variation.
\newblock {\em International Conference on Learning Representations (ICLR)},
  2018.

\bibitem{kholgade20143d}
N.~Kholgade, T.~Simon, A.~Efros, and Y.~Sheikh.
\newblock 3d object manipulation in a single photograph using stock 3d models.
\newblock {\em ACM Transactions on Graphics (TOG)}, 33(4):127, 2014.

\bibitem{kingma2014adam}
D.~P. Kingma and J.~Ba.
\newblock Adam: A method for stochastic optimization.
\newblock In {\em International Conference on Learning Representations (ICLR)},
  2015.

\bibitem{kingma2013auto}
D.~P. Kingma and M.~Welling.
\newblock Auto-encoding variational bayes.
\newblock In {\em International Conference on Learning Representations (ICLR)},
  2014.

\bibitem{lalonde2007photo}
J.-F. Lalonde, D.~Hoiem, A.~A. Efros, C.~Rother, J.~Winn, and A.~Criminisi.
\newblock Photo clip art.
\newblock In {\em ACM transactions on graphics (TOG)}, volume~26, page~3. ACM,
  2007.

\bibitem{lpt2013ikea}
J.~J. Lim, H.~Pirsiavash, and A.~Torralba.
\newblock {Parsing IKEA Objects: Fine Pose Estimation}.
\newblock {\em ICCV}, 2013.

\bibitem{liu2016large}
X.~Liu, W.~Liu, H.~Ma, and H.~Fu.
\newblock Large-scale vehicle re-identification in urban surveillance videos.
\newblock In {\em 2016 IEEE International Conference on Multimedia and Expo
  (ICME)}, pages 1--6. IEEE, 2016.

\bibitem{liu2016deepfashion}
Z.~Liu, P.~Luo, S.~Qiu, X.~Wang, and X.~Tang.
\newblock Deepfashion: Powering robust clothes recognition and retrieval with
  rich annotations.
\newblock In {\em Proceedings of the IEEE conference on CVPR}, pages
  1096--1104, 2016.

\bibitem{lucic2018gans}
M.~Lucic, K.~Kurach, M.~Michalski, S.~Gelly, and O.~Bousquet.
\newblock Are gans created equal? a large-scale study.
\newblock In {\em Advances in neural information processing systems}, pages
  698--707, 2018.

\bibitem{ma2017pose}
L.~Ma, X.~Jia, Q.~Sun, B.~Schiele, T.~Tuytelaars, and L.~Van~Gool.
\newblock Pose guided person image generation.
\newblock In {\em Advances in Neural Information Processing Systems}, pages
  406--416, 2017.

\bibitem{marin2018unsupervised}
P.~A. Mar{\'\i}n-Reyes, L.~Bergamini, J.~Lorenzo-Navarro, A.~Palazzi,
  S.~Calderara, and R.~Cucchiara.
\newblock Unsupervised vehicle re-identification using triplet networks.
\newblock In {\em 2018 IEEE/CVF Conference on CVPR Workshops (CVPRW)}, pages
  166--1665. IEEE, 2018.

\bibitem{mirza2014conditional}
M.~Mirza and S.~Osindero.
\newblock Conditional generative adversarial nets.
\newblock {\em arXiv preprint arXiv:1411.1784}, 2014.

\bibitem{nguyen2019hologan}
T.~Nguyen-Phuoc, C.~Li, L.~Theis, C.~Richardt, and Y.-L. Yang.
\newblock Hologan: Unsupervised learning of 3d representations from natural
  images.
\newblock In {\em Proceedings of the IEEE International Conference on Computer
  Vision}, pages 7588--7597, 2019.

\bibitem{oechsle2019texture}
M.~Oechsle, L.~Mescheder, M.~Niemeyer, T.~Strauss, and A.~Geiger.
\newblock Texture fields: Learning texture representations in function space.
\newblock In {\em Proceedings of the IEEE International Conference on Computer
  Vision}, pages 4531--4540, 2019.

\bibitem{olszewski2019transformable}
K.~Olszewski, S.~Tulyakov, O.~Woodford, H.~Li, and L.~Luo.
\newblock Transformable bottleneck networks.
\newblock In {\em Proceedings of the IEEE International Conference on Computer
  Vision}, pages 7648--7657, 2019.

\bibitem{ortiz2015bayesian}
R.~Ortiz-Cayon, A.~Djelouah, and G.~Drettakis.
\newblock A bayesian approach for selective image-based rendering using
  superpixels.
\newblock In {\em International Conference on 3D Vision-3DV}, 2015.

\bibitem{ortiz2016automatic}
R.~Ortiz-Cayon, A.~Djelouah, F.~Massa, M.~Aubry, and G.~Drettakis.
\newblock Automatic 3d car model alignment for mixed image-based rendering.
\newblock In {\em 2016 Fourth International Conference on 3D Vision (3DV)},
  pages 286--295. IEEE, 2016.

\bibitem{palazzi2018end}
A.~Palazzi, L.~Bergamini, S.~Calderara, and R.~Cucchiara.
\newblock End-to-end 6-dof object pose estimation through differentiable
  rasterization.
\newblock In {\em Second Workshop on 3D Reconstruction Meets Semantics
  (3DRMS)}, 2018.

\bibitem{park2017transformation}
E.~Park, J.~Yang, E.~Yumer, D.~Ceylan, and A.~C. Berg.
\newblock Transformation-grounded image generation network for novel 3d view
  synthesis.
\newblock In {\em 2017 IEEE Conference on Computer Vision and Pattern
  Recognition (CVPR)}, pages 702--711. IEEE, 2017.

\bibitem{paszke2017pytorch}
A.~Paszke, S.~Gross, S.~Chintala, G.~Chanan, E.~Yang, Z.~DeVito, Z.~Lin,
  A.~Desmaison, L.~Antiga, and A.~Lerer.
\newblock Automatic differentiation in pytorch.
\newblock 2017.

\bibitem{pavlakos20176}
G.~Pavlakos, X.~Zhou, A.~Chan, K.~G. Derpanis, and K.~Daniilidis.
\newblock 6-dof object pose from semantic keypoints.
\newblock In {\em 2017 IEEE International Conference on Robotics and Automation
  (ICRA)}, pages 2011--2018. IEEE, 2017.

\bibitem{qi2018semi}
X.~Qi, Q.~Chen, J.~Jia, and V.~Koltun.
\newblock Semi-parametric image synthesis.
\newblock In {\em Proceedings of the IEEE Conference on CVPR}, pages
  8808--8816, 2018.

\bibitem{radford2015unsupervised}
A.~Radford, L.~Metz, and S.~Chintala.
\newblock Unsupervised representation learning with deep convolutional
  generative adversarial networks.
\newblock {\em International Conference on Learning Representations (ICLR)},
  2016.

\bibitem{rematas2017novel}
K.~Rematas, C.~H. Nguyen, T.~Ritschel, M.~Fritz, and T.~Tuytelaars.
\newblock Novel views of objects from a single image.
\newblock {\em IEEE transactions on pattern analysis and machine intelligence},
  39(8):1576--1590, 2017.

\bibitem{ronneberger2015unet}
O.~Ronneberger, P.~Fischer, and T.~Brox.
\newblock U-net: Convolutional networks for biomedical image segmentation.
\newblock In {\em International Conference on Medical image computing and
  computer-assisted intervention}, pages 234--241. Springer, 2015.

\bibitem{shen2017learning}
Y.~Shen, T.~Xiao, H.~Li, S.~Yi, and X.~Wang.
\newblock Learning deep neural networks for vehicle re-id with
  visual-spatio-temporal path proposals.
\newblock In {\em Proceedings of the IEEE International Conference on Computer
  Vision}, pages 1900--1909, 2017.

\bibitem{shepard1971mental}
R.~N. Shepard and J.~Metzler.
\newblock Mental rotation of three-dimensional objects.
\newblock {\em Science}, 171(3972):701--703, 1971.

\bibitem{shilane2004princeton}
P.~Shilane, P.~Min, M.~Kazhdan, and T.~Funkhouser.
\newblock The princeton shape benchmark.
\newblock In {\em Proceedings Shape Modeling Applications, 2004.}, pages
  167--178. IEEE, 2004.

\bibitem{siarohin2018deformable}
A.~Siarohin, E.~Sangineto, S.~Lathuili{\`e}re, and N.~Sebe.
\newblock Deformable gans for pose-based human image generation.
\newblock In {\em CVPR 2018-Computer Vision and Pattern Recognition}, 2018.

\bibitem{simonyan2014very}
K.~Simonyan and A.~Zisserman.
\newblock Very deep convolutional networks for large-scale image recognition.
\newblock {\em arXiv preprint arXiv:1409.1556}, 2014.

\bibitem{sinha2017surfnet}
A.~Sinha, A.~Unmesh, Q.~Huang, and K.~Ramani.
\newblock Surfnet: Generating 3d shape surfaces using deep residual networks.
\newblock In {\em IEEE CVPR}, 2017.

\bibitem{sitzmann2019deepvoxels}
V.~Sitzmann, J.~Thies, F.~Heide, M.~Nie{\ss}ner, G.~Wetzstein, and
  M.~Zollhofer.
\newblock Deepvoxels: Learning persistent 3d feature embeddings.
\newblock In {\em Proceedings of the IEEE Conference on CVPR}, pages
  2437--2446, 2019.

\bibitem{sun2018multi}
S.-H. Sun, M.~Huh, Y.-H. Liao, N.~Zhang, and J.~J. Lim.
\newblock Multi-view to novel view: Synthesizing novel views with self-learned
  confidence.
\newblock In {\em Proceedings of the European Conference on Computer Vision
  (ECCV)}, pages 155--171, 2018.

\bibitem{sun2018pix3d}
X.~Sun, J.~Wu, X.~Zhang, Z.~Zhang, C.~Zhang, T.~Xue, J.~B. Tenenbaum, and W.~T.
  Freeman.
\newblock Pix3d: Dataset and methods for single-image 3d shape modeling.
\newblock In {\em Proceedings of the IEEE Conference on CVPR}, pages
  2974--2983, 2018.

\bibitem{tatarchenko2016multi}
M.~Tatarchenko, A.~Dosovitskiy, and T.~Brox.
\newblock Multi-view 3d models from single images with a convolutional network.
\newblock In {\em European Conference on Computer Vision}, pages 322--337.
  Springer, 2016.

\bibitem{teichmann2018multinet}
M.~Teichmann, M.~Weber, M.~Zoellner, R.~Cipolla, and R.~Urtasun.
\newblock Multinet: Real-time joint semantic reasoning for autonomous driving.
\newblock In {\em 2018 IEEE Intelligent Vehicles Symposium (IV)}, pages
  1013--1020. IEEE, 2018.

\bibitem{tulsiani2015viewpoints}
S.~Tulsiani and J.~Malik.
\newblock Viewpoints and keypoints.
\newblock In {\em Proceedings of the IEEE Conference on CVPR}, pages
  1510--1519, 2015.

\bibitem{wang2018high}
T.-C. Wang, M.-Y. Liu, J.-Y. Zhu, A.~Tao, J.~Kautz, and B.~Catanzaro.
\newblock High-resolution image synthesis and semantic manipulation with
  conditional gans.
\newblock In {\em Proceedings of the IEEE Conference on CVPR}, pages
  8798--8807, 2018.

\bibitem{wang2017veri}
Z.~Wang, L.~Tang, X.~Liu, Z.~Yao, S.~Yi, J.~Shao, J.~Yan, S.~Wang, H.~Li, and
  X.~Wang.
\newblock Orientation invariant feature embedding and spatial temporal
  regularization for vehicle re-identification.
\newblock In {\em The IEEE International Conference on Computer Vision (ICCV)},
  Oct 2017.

\bibitem{wu2017marrnet}
J.~Wu, Y.~Wang, T.~Xue, X.~Sun, B.~Freeman, and J.~Tenenbaum.
\newblock Marrnet: 3d shape reconstruction via 2.5 d sketches.
\newblock In {\em Advances in neural information processing systems}, pages
  540--550, 2017.

\bibitem{wu20183d}
J.~Wu, T.~Xue, J.~J. Lim, Y.~Tian, J.~B. Tenenbaum, A.~Torralba, and W.~T.
  Freeman.
\newblock 3d interpreter networks for viewer-centered wireframe modeling.
\newblock {\em International Journal of Computer Vision}, 126(9):1009--1026,
  2018.

\bibitem{xian2018texturegan}
W.~Xian, P.~Sangkloy, V.~Agrawal, A.~Raj, J.~Lu, C.~Fang, F.~Yu, and J.~Hays.
\newblock Texturegan: Controlling deep image synthesis with texture patches.
\newblock In {\em Proceedings of the IEEE Conference on CVPR}, pages
  8456--8465, 2018.

\bibitem{xiang2016objectnet3d}
Y.~Xiang, W.~Kim, W.~Chen, J.~Ji, C.~Choy, H.~Su, R.~Mottaghi, L.~Guibas, and
  S.~Savarese.
\newblock Objectnet3d: A large scale database for 3d object recognition.
\newblock In {\em European Conference on Computer Vision}, pages 160--176.
  Springer, 2016.

\bibitem{xiang2014beyond}
Y.~Xiang, R.~Mottaghi, and S.~Savarese.
\newblock Beyond pascal: A benchmark for 3d object detection in the wild.
\newblock In {\em Applications of Computer Vision (WACV), 2014 IEEE Winter
  Conference on}, pages 75--82. IEEE, 2014.

\bibitem{xiao2019identity}
F.~Xiao, H.~Liu, and Y.~J. Lee.
\newblock Identity from here, pose from there: Self-supervised disentanglement
  and generation of objects using unlabeled videos.
\newblock In {\em Proceedings of the IEEE International Conference on Computer
  Vision}, pages 7013--7022, 2019.

\bibitem{yang2015weakly}
J.~Yang, S.~E. Reed, M.-H. Yang, and H.~Lee.
\newblock Weakly-supervised disentangling with recurrent transformations for 3d
  view synthesis.
\newblock In {\em Advances in Neural Information Processing Systems}, pages
  1099--1107, 2015.

\bibitem{yang2018denseaspp}
M.~Yang, K.~Yu, C.~Zhang, Z.~Li, and K.~Yang.
\newblock Denseaspp for semantic segmentation in street scenes.
\newblock In {\em Proceedings of the IEEE Conference on CVPR}, pages
  3684--3692, 2018.

\bibitem{yu2018generative}
J.~Yu, Z.~Lin, J.~Yang, X.~Shen, X.~Lu, and T.~S. Huang.
\newblock Generative image inpainting with contextual attention.
\newblock {\em Proceedings of the IEEE Conference on Computer Vision and
  Pattern Recognition}, 2018.

\bibitem{zhang2017stackgan}
H.~Zhang, T.~Xu, H.~Li, S.~Zhang, X.~Wang, X.~Huang, and D.~N. Metaxas.
\newblock Stackgan: Text to photo-realistic image synthesis with stacked
  generative adversarial networks.
\newblock In {\em Proceedings of the IEEE International Conference on Computer
  Vision}, pages 5907--5915, 2017.

\bibitem{zhao2018multi}
B.~Zhao, X.~Wu, Z.-Q. Cheng, H.~Liu, Z.~Jie, and J.~Feng.
\newblock Multi-view image generation from a single-view.
\newblock In {\em 2018 ACM Multimedia Conference on Multimedia Conference},
  pages 383--391. ACM, 2018.

\bibitem{zheng2015scalable}
L.~Zheng, L.~Shen, L.~Tian, S.~Wang, J.~Wang, and Q.~Tian.
\newblock Scalable person re-identification: A benchmark.
\newblock In {\em Computer Vision, IEEE International Conference on}, 2015.

\bibitem{zhou2018open3d}
Q.-Y. Zhou, J.~Park, and V.~Koltun.
\newblock {Open3D}: {A} modern library for {3D} data processing.
\newblock {\em arXiv:1801.09847}, 2018.

\bibitem{zhou2016view}
T.~Zhou, S.~Tulsiani, W.~Sun, J.~Malik, and A.~A. Efros.
\newblock View synthesis by appearance flow.
\newblock In {\em European conference on computer vision}, pages 286--301.
  Springer, 2016.

\bibitem{zhu2017unpaired}
J.-Y. Zhu, T.~Park, P.~Isola, and A.~A. Efros.
\newblock Unpaired image-to-image translation using cycle-consistent
  adversarial networks.
\newblock {\em IEEE International Conference on Computer Vision (ICCV)}, 2017.

\bibitem{zhu2017toward}
J.-Y. Zhu, R.~Zhang, D.~Pathak, T.~Darrell, A.~A. Efros, O.~Wang, and
  E.~Shechtman.
\newblock Toward multimodal image-to-image translation.
\newblock In {\em Advances in Neural Information Processing Systems}, pages
  465--476, 2017.

\bibitem{zhu2018visual}
J.-Y. Zhu, Z.~Zhang, C.~Zhang, J.~Wu, A.~Torralba, J.~Tenenbaum, and
  B.~Freeman.
\newblock Visual object networks: image generation with disentangled 3d
  representations.
\newblock In {\em Advances in Neural Information Processing Systems}, pages
  118--129, 2018.

\end{thebibliography}
}

%
%
\begin{IEEEbiography}[{\includegraphics[width=1in,height=1.25in,clip,keepaspectratio]{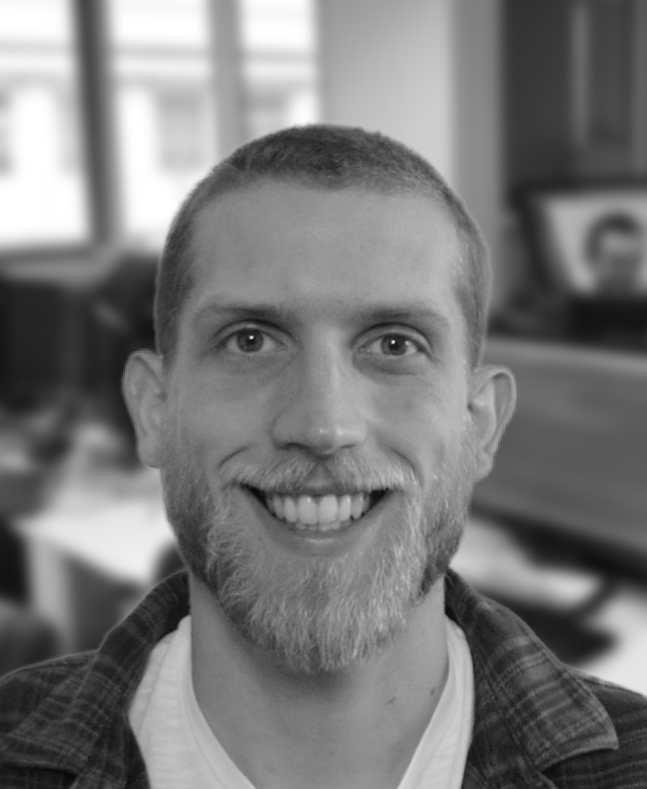}}]{Andrea Palazzi}
MSc in computer engineering in 2015, PhD in 2019 at University of Modena and Reggio Emilia, Italy. His PhD research revolved around the topics of driver’s attention prediction and gaze estimation, use of synthetic data to train deep neural networks, synthesis of novel views from single monocular images. He is currently Senior Deep Learning Engineer at Nomitri, where he tackles the challenge of bringing visual intelligence to smartphones and edge devices.
\end{IEEEbiography}
%
\begin{IEEEbiography}[{\includegraphics[width=1in,height=1.25in,clip,keepaspectratio]{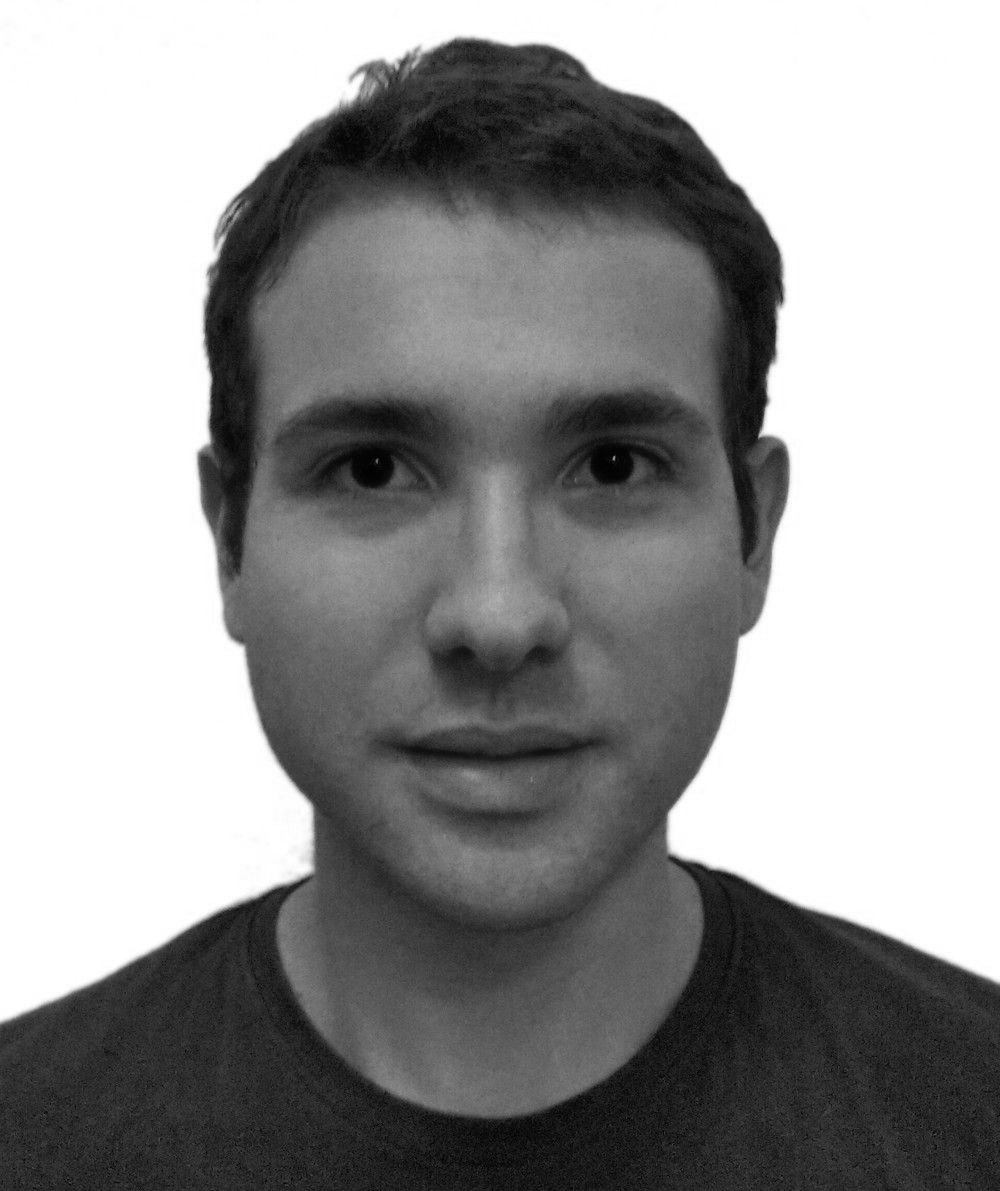}}]{Luca Bergamini}
is currently in his last year of PhD in the AImageLab at the University of Modena and Reggio Emilia, where he received his master degree in 2017. His research interests lie in the area of computer vision and deep learning. He has recently joined Lyft as a research SWE.
\end{IEEEbiography}
\begin{IEEEbiography}[{\includegraphics[width=1in,height=1.25in,clip,keepaspectratio]{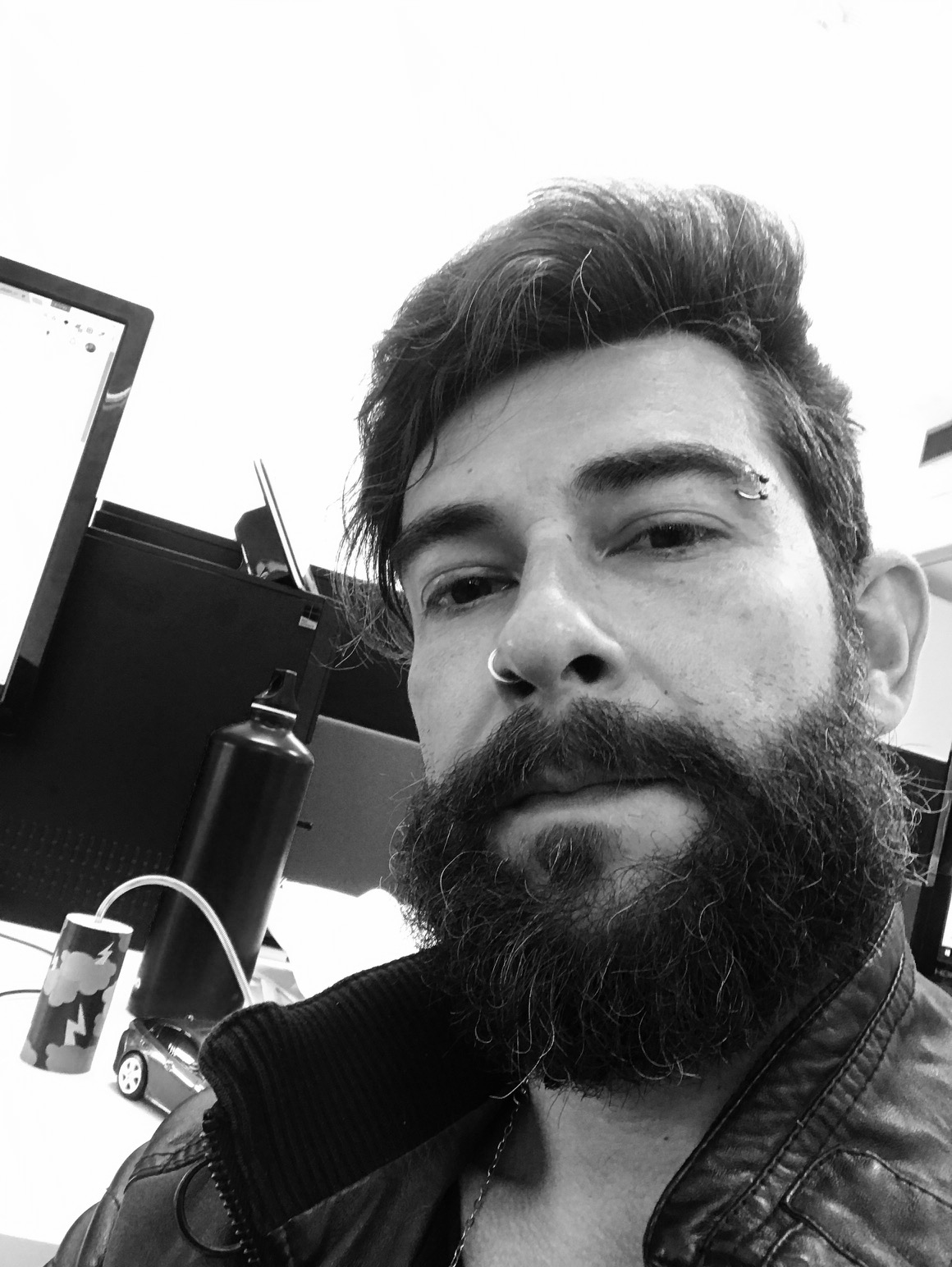}}]{Simone Calderara}
received a computer engineering master's degree in 2005 and the PhD degree in 2009 from the University of Modena and Reggio Emilia, where he is currently an assistant professor within the Imagelab group. His current research interests include computer vision and machine learning applied to human behavior analysis, visual tracking in crowded scenarios, and time series analysis for forensic applications. He is a member of the IEEE.
\end{IEEEbiography}
\begin{IEEEbiography}[{\includegraphics[width=1in,height=1.25in,clip,keepaspectratio]{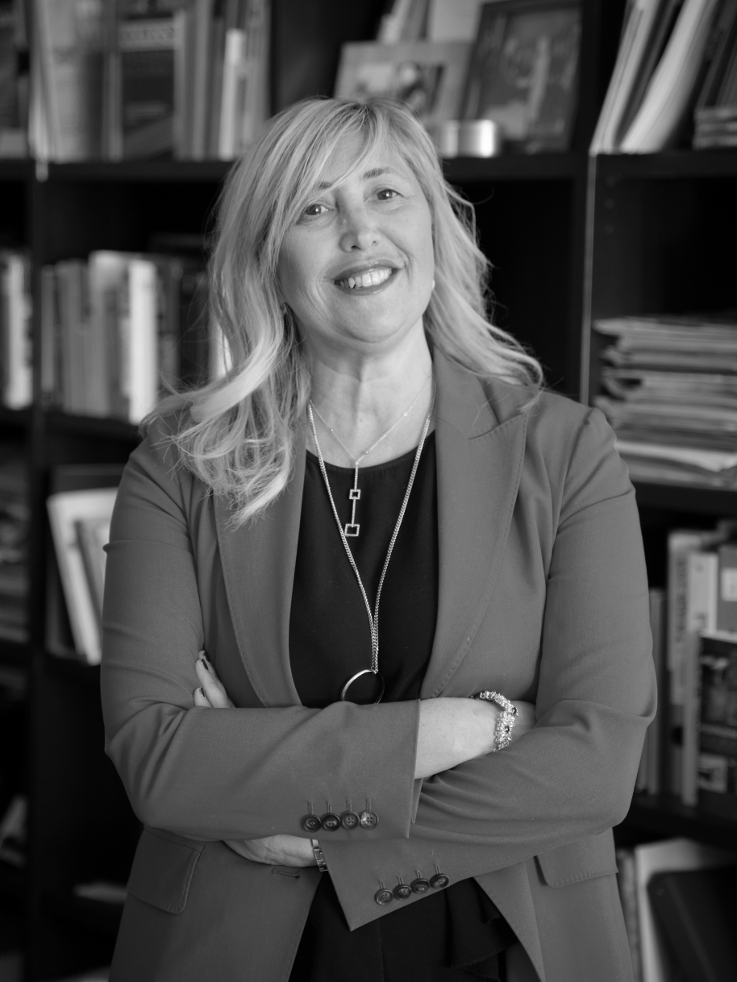}}]{Rita Cucchiara}
(PhD in Computer Engineering 1992, University of Bologna, Italy) is full professor at the University of Modena and Reggio Emilia, Italy, where she heads the AImageLab group.
Director of the Italian Lab CINI Of Artificial Intelligence and intelligent Systems. Former President of the Italian CVPL association. IAPR Fellow since 2006, ELLIS fellow since 2019. Part of the Governing Board of IAPR and a member of the the Advisory Board of CVF. Area, Program, General Chair of several top-tier conferences. She published more than 400 papers on pattern recognition, computer vision and multimedia.
\end{IEEEbiography}


\clearpage
\twocolumn[
   \begin{center}
      \huge\textbf{SUPPLEMENTARY MATERIAL\\Warp and Learn: Novel Views Generation for Vehicles and Other Objects}\\
   \end{center}
]
\noindent Here we provide additional material useful for the understanding of the paper. We remind the reader that code, data and animated results are available at: \href{https://github.com/ndrplz/semiparametric}{https://github.com/ndrplz/semiparametric}

%
\begin{figure*}[th]
    \centering
    \includegraphics[width=0.98\textwidth]{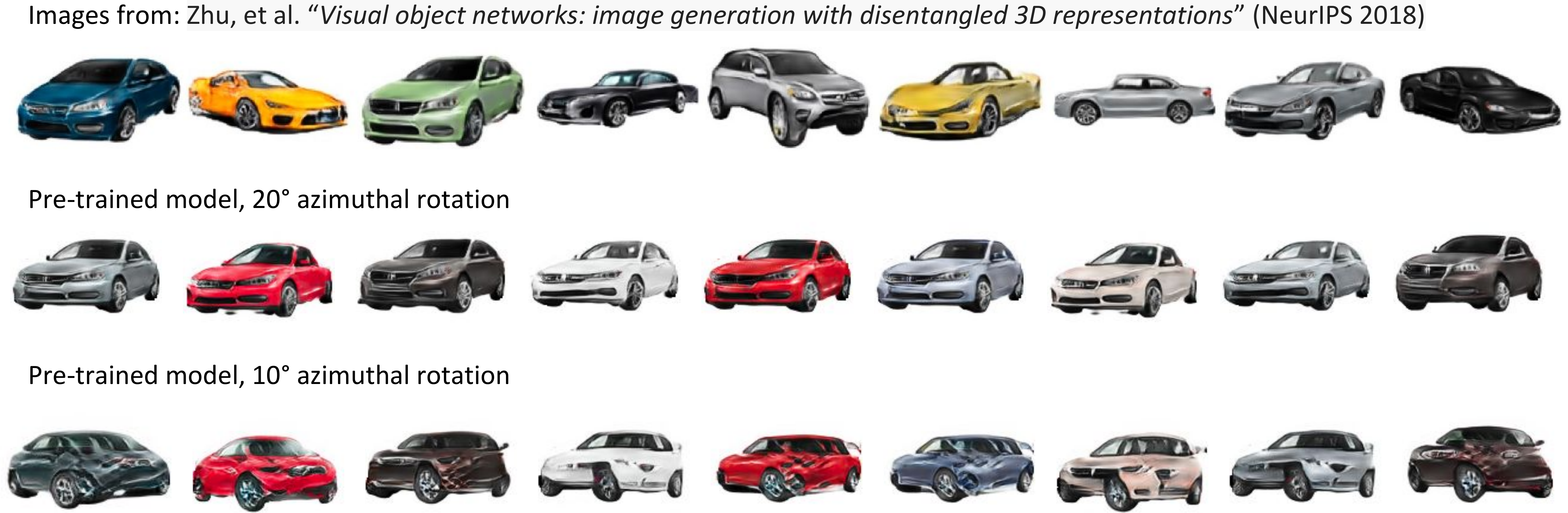}
    \caption{Qualitative results from VON~\cite{zhu2018visual}. Since the model has no geometric constraints, a very small viewpoint variation from a situation where it performs impressively (middle) can lead to a dramatic failure (bottom). Please see Sec.~\ref{sec:suppl:discussion} for details.}
    \label{fig:suppl:von_reinvent_wheel}
\end{figure*}
\begin{figure*}[t]
    \centering
    \includegraphics[width=0.98\textwidth]{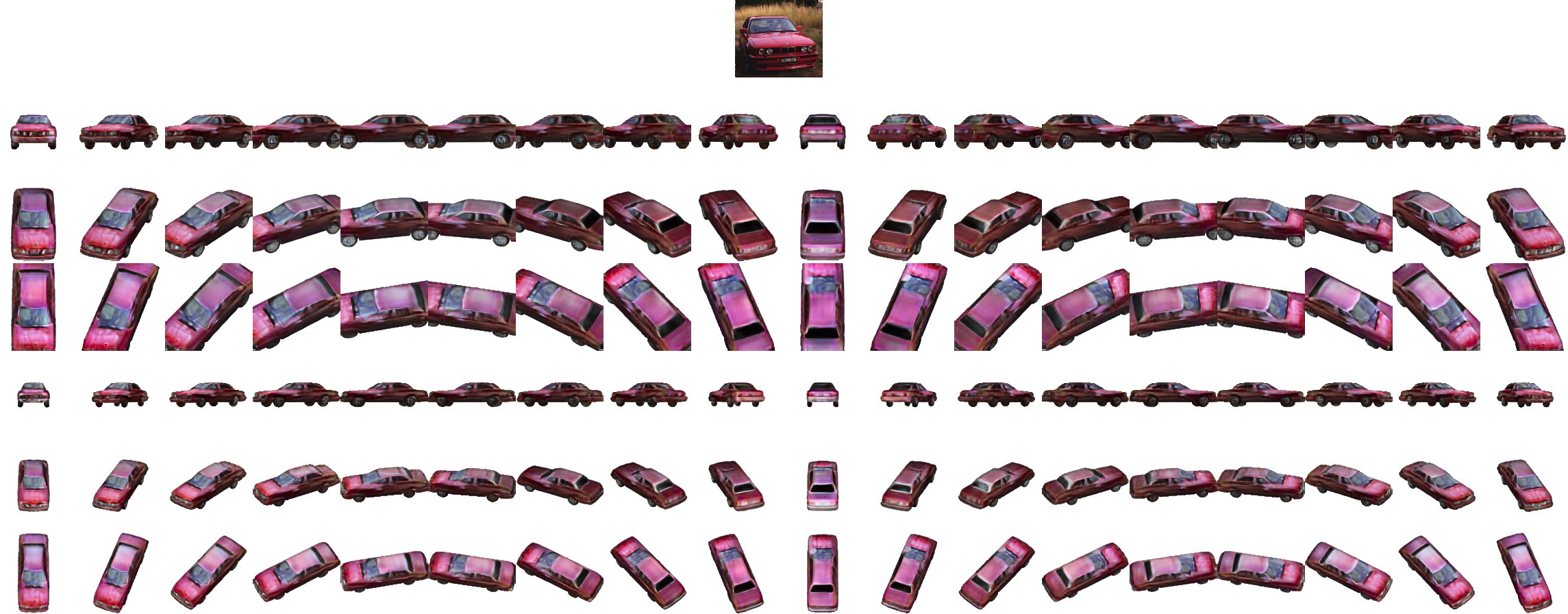}
    \caption{The the explicit texture warping and the geometric guidance given by the 2.5 sketches allows our method to produce consistent predictions from arbitrary viewpoint even tough no explicit consistency loss between different views is optimized. First row shows input. Best viewed in color.}
    \label{fig:suppl:output_consistency}
\end{figure*}
\begin{figure*}[th!]
    \centering
    \includegraphics[width=\textwidth]{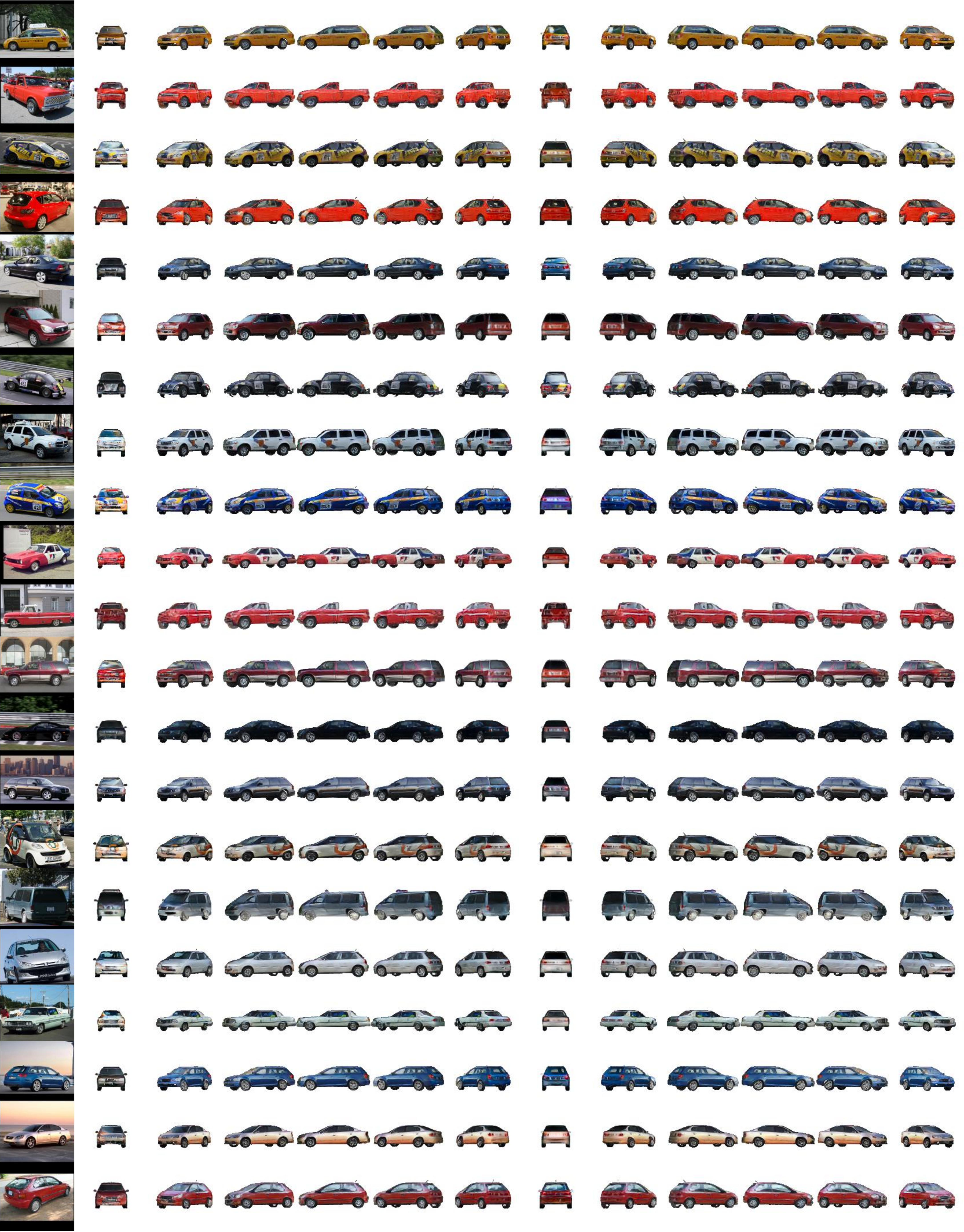}
    \caption{Visual examples of 360 degrees car rotation. Best viewed zoomed on screen.}
    \label{fig:suppl:rot_cars}
\end{figure*}
\begin{figure*}[th!]
    \centering
    \includegraphics[width=0.95\textwidth]{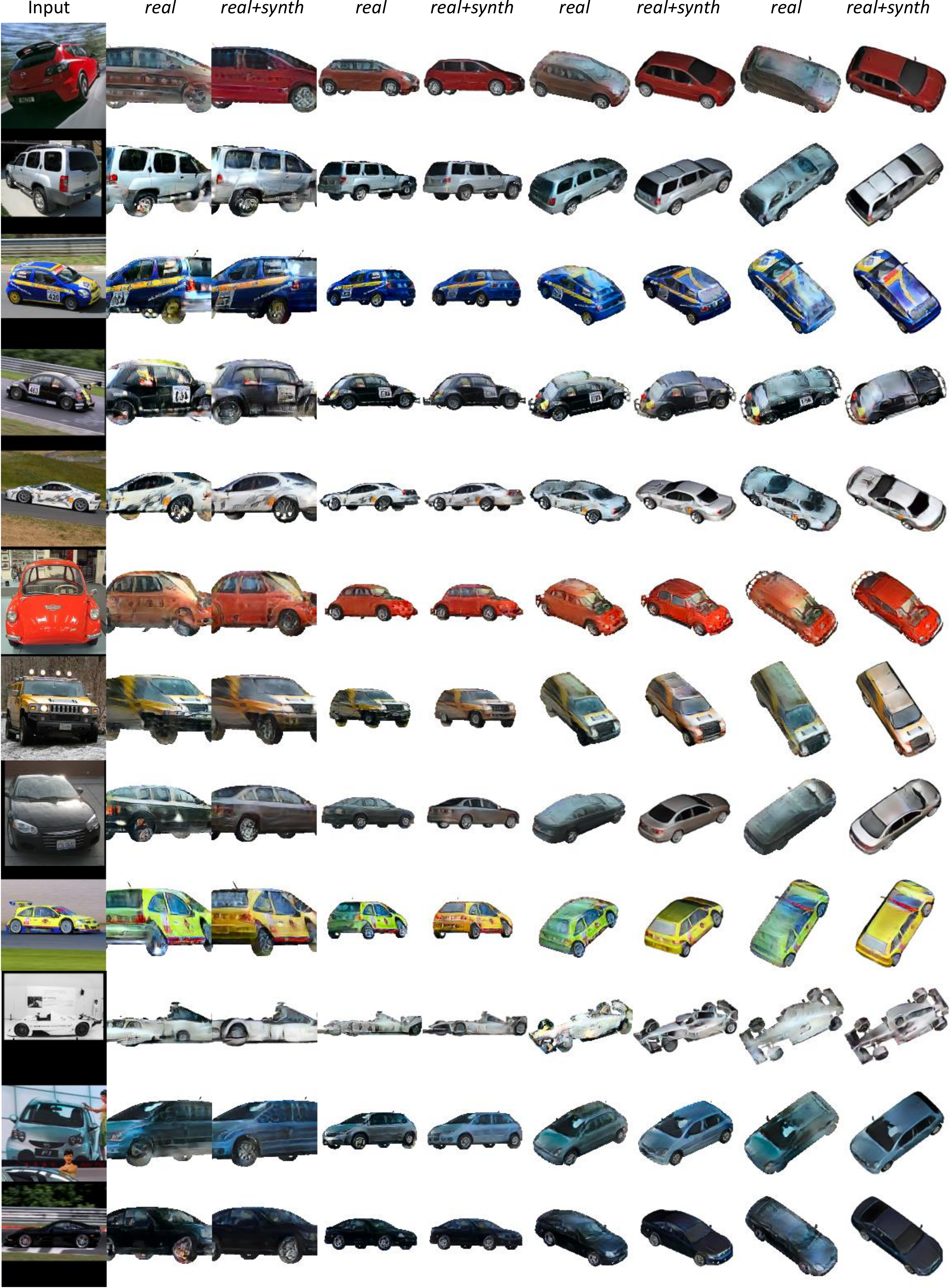}
    \caption{Additional visual results from our model, showing how training the \icn on a mix of synthetic and real data dramatically improves the performance for out-of-distribution viewpoints.}
    \label{fig:suppl:real_vs_synth}
\end{figure*}
\begin{figure*}[th!]
    \centering
    \includegraphics[width=\textwidth]{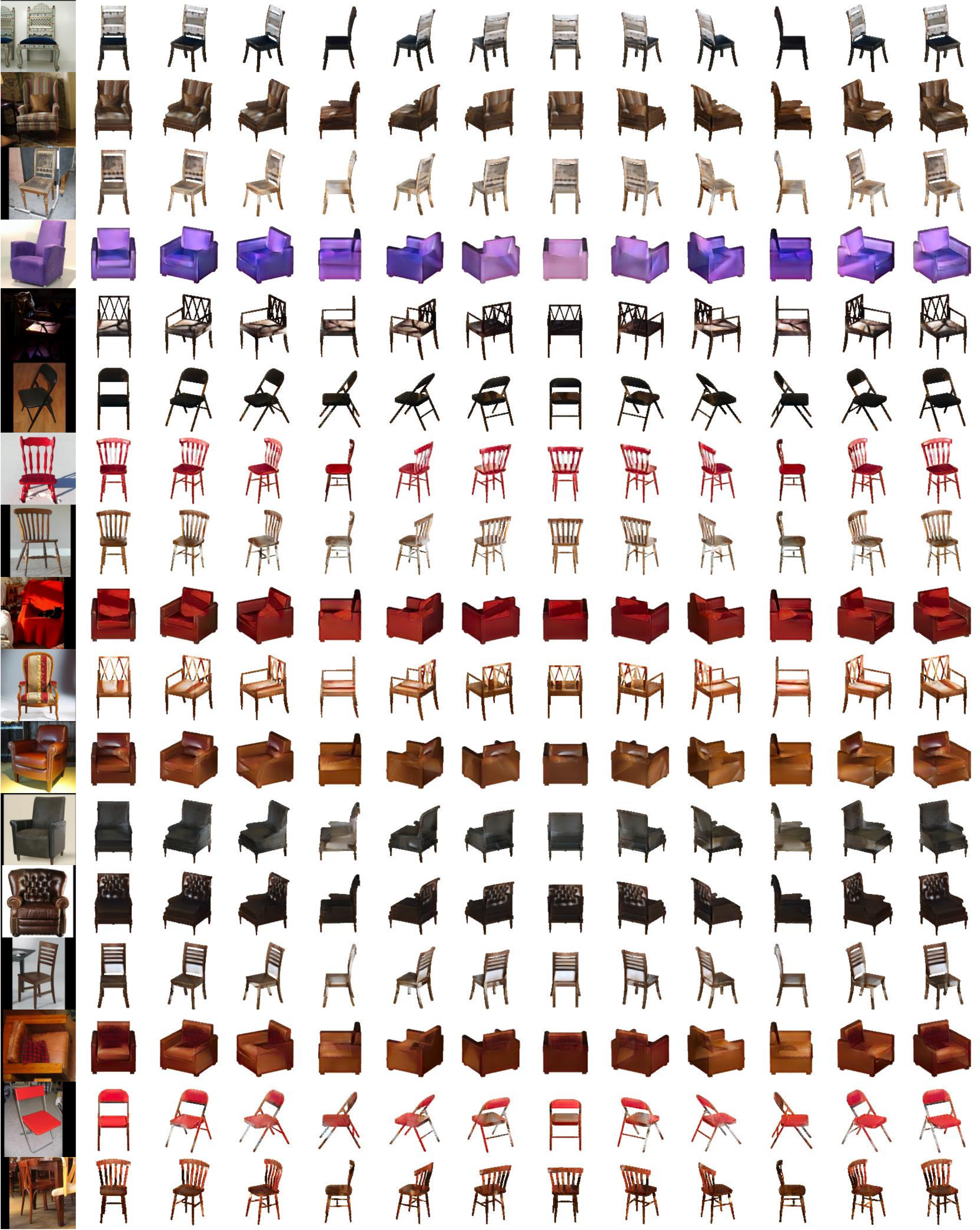}
    \caption{Visual examples of 360 degrees chair rotation. Best viewed zoomed on screen.}
    \label{fig:suppl:rot_chairs}
\end{figure*}
\begin{figure*}[th!]
    \centering
    \includegraphics[width=0.95\textwidth]{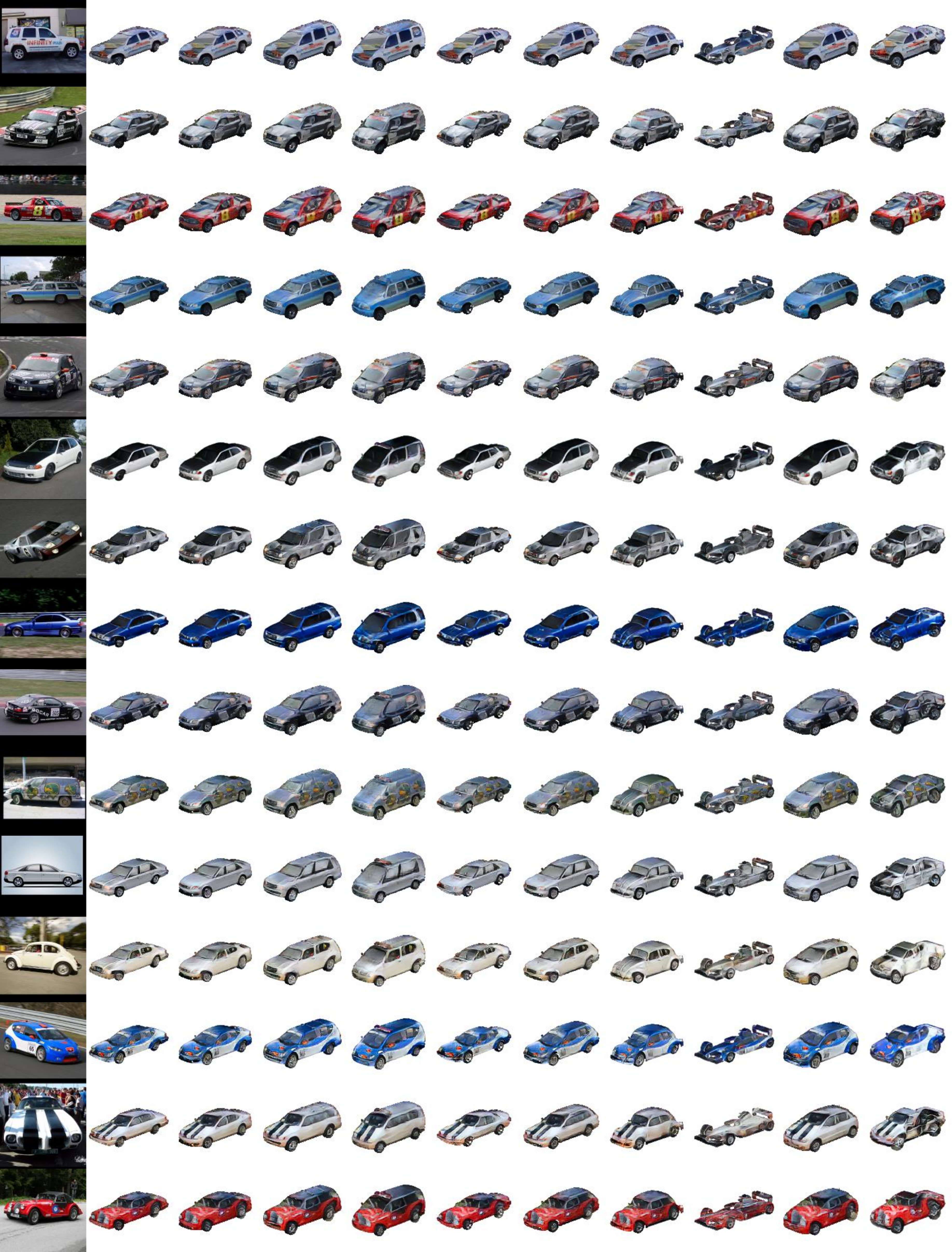}
    \caption{Additional visual results for shape transfer on vehicle class.}
    \label{fig:suppl:car_shape_transfer}
\end{figure*}
\section{Discussion on output consistency}
\label{sec:suppl:discussion}
In this work we argue that a semi-parametric method for object novel view synthesis has very appealing properties coming from the fact that the output is not generated from scratch; on the opposite, much of the information is warped from the input viewpoint in a geometrically-principled manner. We  find that one of the major advantages of this semi-parametric approach is that the geometric guidance helps avoiding catastrophic failures in the generation process.\\ \\
In particular, our model does not have to explicitly learn the hard concept of distance between two viewpoints in 3D space. Since the 2.5D sketch which guide the generation process are rendered from a 3D model in a purely geometric fashion, continuous change in viewpoint will lead by construction to continuous variation in the 2.5D sketch.\\ \\
Conversely, in a completely learning-based pipeline the model is let to learn the concept of 3D viewpoint proximity from a massive amount of data. Still, it is a very hard concept: the model may or may not learn it properly. In Fig.~\ref{fig:suppl:von_reinvent_wheel} we make use of the VON~\cite{zhu2018visual} pre-trained model to showcase one situation of this kind. Even though the model is state-of-the-art and its performance is impressive, very little viewpoint variation can result in a dramatic failure in the image generation for no apparent reason.\\
On the opposite, the explicit texture warping and the geometric guidance given by the 2.5 sketch allows our method to produce consistent predictions from arbitrary viewpoints even tough no explicit consistency loss between different views is optimized. Visual examples from our methods are shown in Figures~\ref{fig:suppl:output_consistency},\ref{fig:suppl:real_vs_synth},\ref{fig:suppl:rot_cars},\ref{fig:suppl:rot_chairs}, where we drastically jointly change both azimuth and elevation of the target viewpoint.\\ \\
Additional visual results for shape transfer on the vehicle class are depicted in Fig.~\ref{fig:suppl:car_shape_transfer}.
%
%
\section{Keypoints details}
\label{sec:suppl:keypoints}
\begin{figure*}[t]
    \centering
    \includegraphics[width=\textwidth]{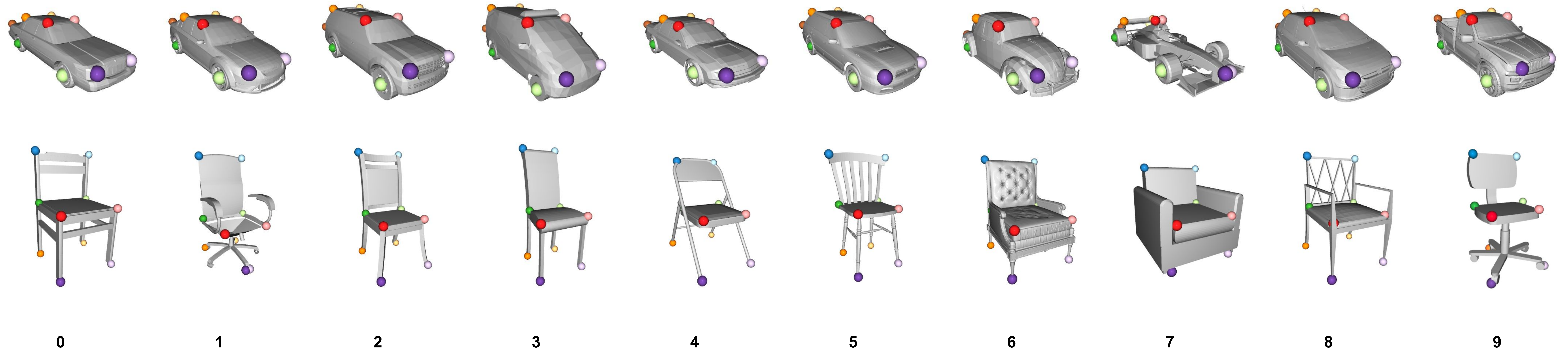}
    \caption{Semantic keypoints annotated for each 3D model in the Pascal3D+~\cite{xiang2014beyond} dataset: bottom row indicates the CAD index in the dataset. These are the same keypoints we choose to annotate in our synthetic dataset. Please refer to Sec.~\ref{sec:suppl:keypoints} for further discussion.}
    \label{fig:suppl:pascal_cads}
\end{figure*}
\begin{figure}[t]
    \centering
    \includegraphics[width=0.9\columnwidth]{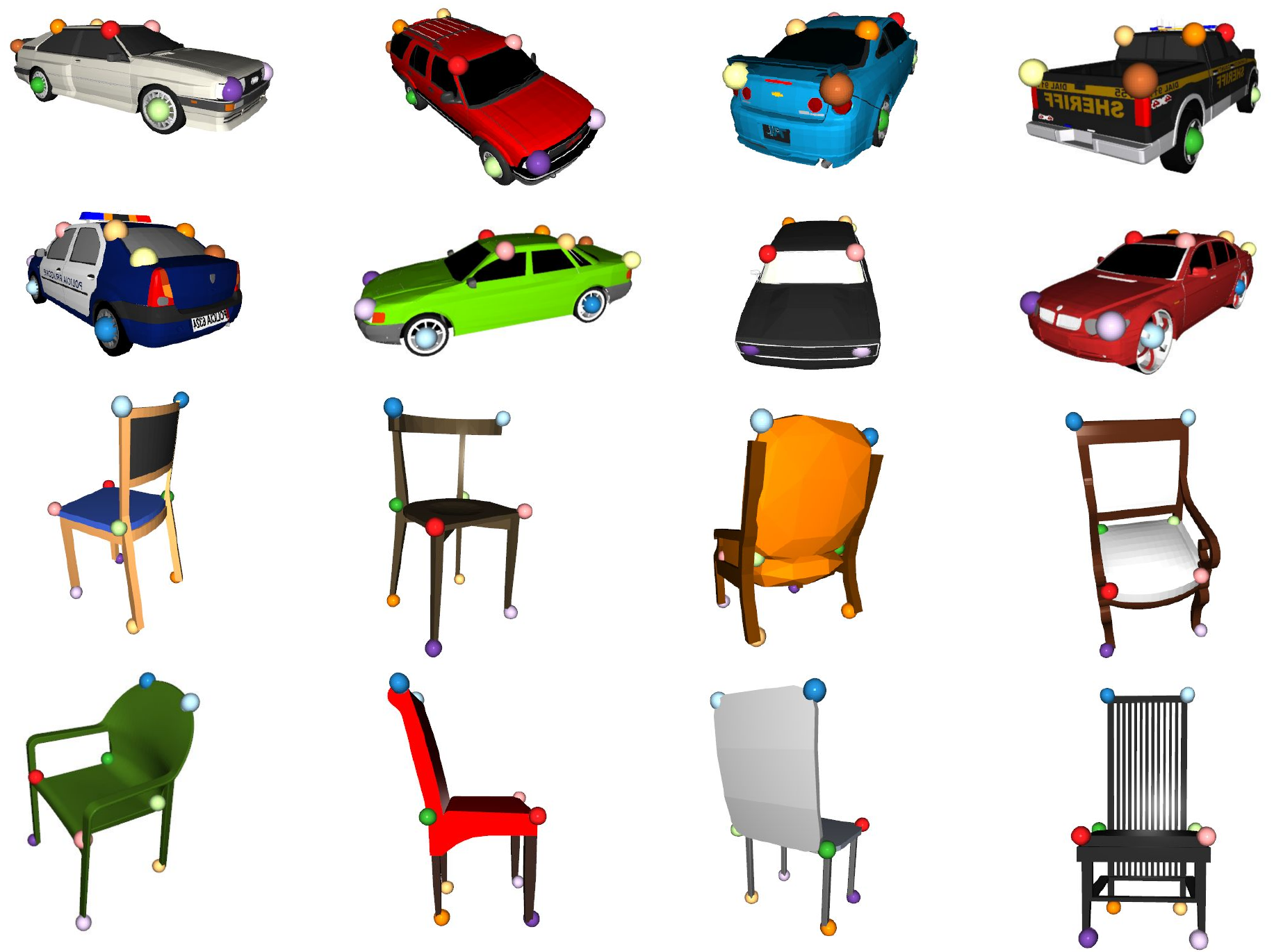}
    \caption{Random 3D models from our synthetic dataset, rendered together with their annotated 3D keypoints.}
    \label{fig:synth_ex}
\end{figure}

In this work semantic keypoints are leveraged as a proxy to decompose the object in the image into a small set of planar faces, with the keypoints constituting the vertices of each face.\\
There is no universal agreement about the number of significant objects' landmarks, not even for most common classes such as vehicles. Indeed, different works have often used a different number of keypoints to characterize the same objects~\cite{lpt2013ikea,xiang2014beyond,xiang2016objectnet3d,wang2017veri,wu20183d,huang2018apolloscape}.\\ \\
Here we follow the convention of the \pascal~\cite{xiang2014beyond} dataset, which defines 12 keypoints for vehicles (front and back wheels, upper windshield, upper rear window, front light, back trunk - left and right) and 10 for chairs (back upper, seat upper, seat lower, leg upper, leg lower - left and right).\\
The location of these keypoints on the \pascal models can be visually inspected in Fig.~\ref{fig:suppl:pascal_cads}; some examples of annotated models are depicted in Fig.~\ref{fig:synth_ex}.
%
\section{Planes details}
\label{sec:suppl:planes}
Our method relies on a set of planes delimited by 2D keypoints, where each plane is defined by at least 3 keypoints. The size of this set is an hyper-parameter for our model. In fact, this value must be set accordingly to balance two factors. The first one is that, as the number of planes increases the same must hold true for the number of available keypoints. Ideally, when the number of keypoints reaches infinity, the set of planes contains the 3D surface of the object. However, the number of annotated keypoints in a dataset is seldom higher than a few dozens. On the other hand, when the number of keypoints is extremely low the set of planes hardly approximate the surface of the object. However, we found that six planes are enough for the car setting, while four suffices for the chair one.

We believe this to be due to the presence of textures and other high-frequency details on these planes. Although the side of a vehicle is not a single flat surface (thus introducing a strong approximation while relying on planar homography for the warping) the presence of those details tricks the human's eye in believing the surface has in fact three dimensions. Then, the \icn is trained to fix inconsistencies at the borders between warped planes and to merge all the patches into a seamless figure.
%
\section{CAD alignment details}
\begin{figure}[t]
    \centering
    \includegraphics[width=0.95\columnwidth]{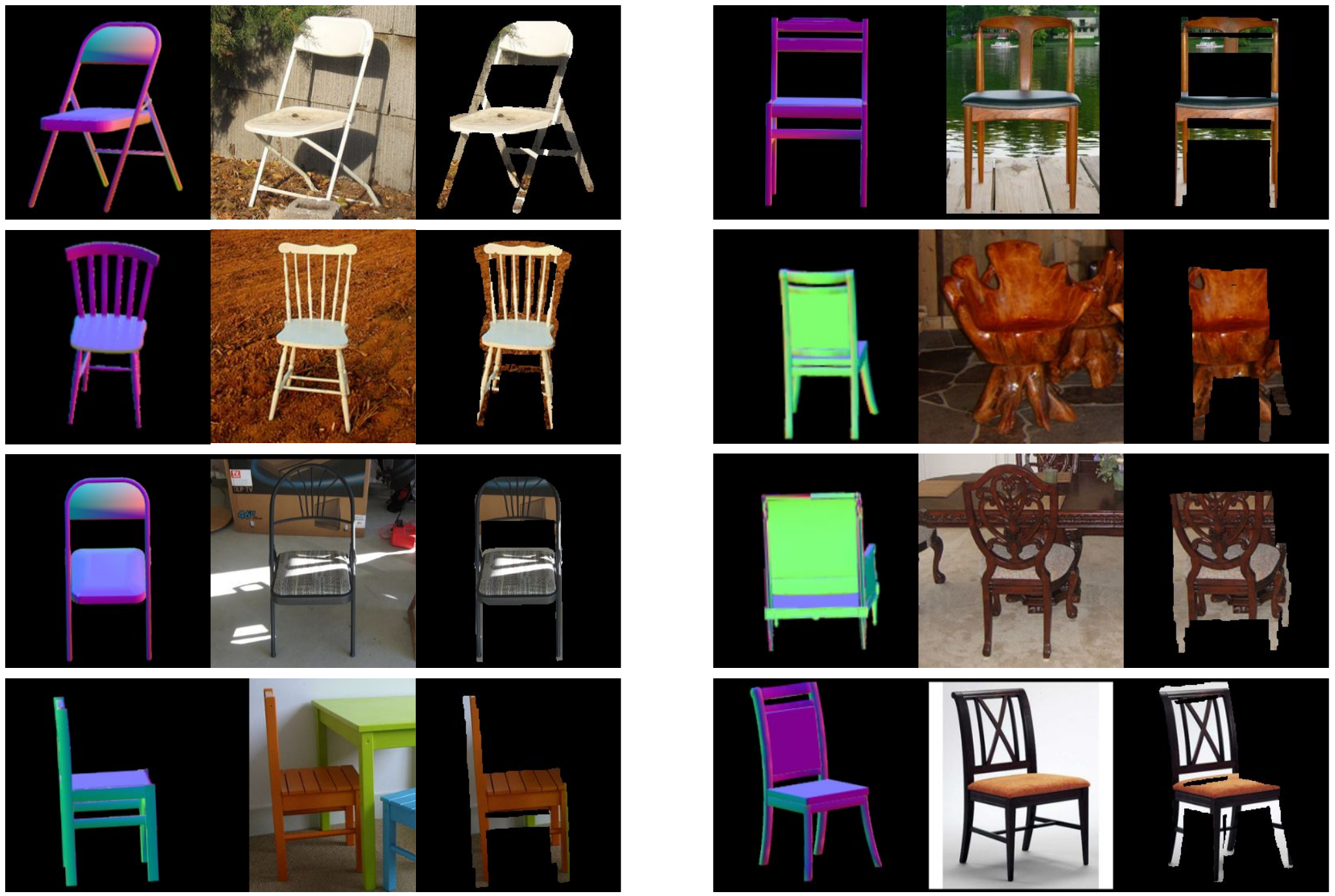}
    \caption{Examples of misalignment between the source image and the annotated cad model. Mainly sources of error consist of wrong annotated model viewpoint and the model not resembling the depicted object.}
    \label{fig:chairs_align}
\end{figure}

As mentioned in experimental section of the main paper, CAD models are not perfectly aligned with images in \pascal dataset. This is due to two main reasons. Firstly, the CAD models have been placed by human workers and then refined using an automatic algorithm. Several geometric simplifications and assumptions are introduces (e.g. about camera intrinsic), and re-projection errors and human mistakes may arise. Accordingly to \pascal~\cite{xiang2014beyond}:
\begin{quote}
    The annotator (...) rotates the 3D CAD model (...). The alignment provides us with rough azimuth and elevation angles, which are used as initialization (...). We assume a simplified camera model, where the world coordinate is defined on the 3D CAD model and the camera is facing the origin of the world coordinate system (...).
\end{quote}
Secondly, those CAD models don't match the object portrayed, as they hail from a very reduced subset of only 10 elements. Again, from \pascal~\cite{xiang2014beyond}:
\begin{quote}
The annotator first selects the 3D CAD model that best resembles the object instance (...).
\end{quote}
Figure \ref{fig:chairs_align} shows some example of the effects of this misalignment for the chair subset, where holes and slits present an additional source of error.
\label{sec:suppl:cad}
%
%
\section{Failure cases}
\label{sec:suppl:failure_cases}
\begin{figure*}[b]
    \centering
    \includegraphics[width=0.95\textwidth]{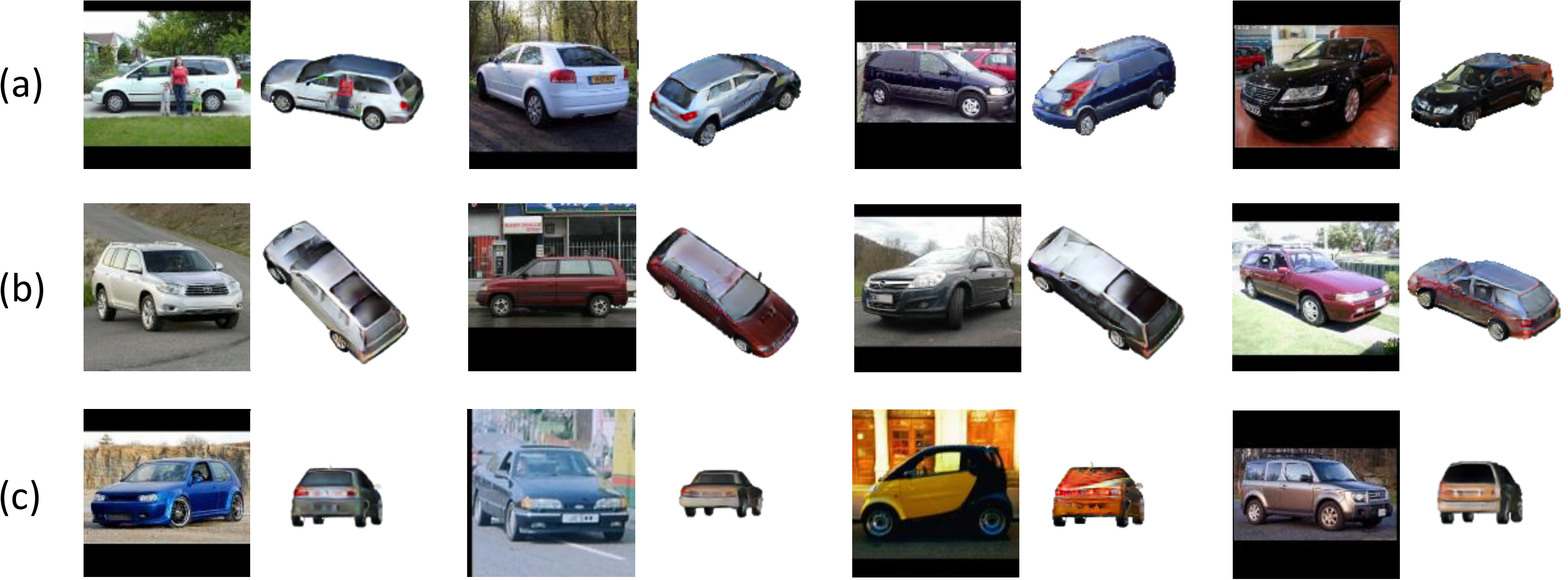}
    \caption{Most common failure cases of our model on the vehicle class. Please see Sec.~\ref{sec:suppl:failure_cases} for details.}
    \label{fig:failure_cases_cars}
\end{figure*}
\begin{figure*}[t]
    \centering
    \includegraphics[width=0.95\textwidth]{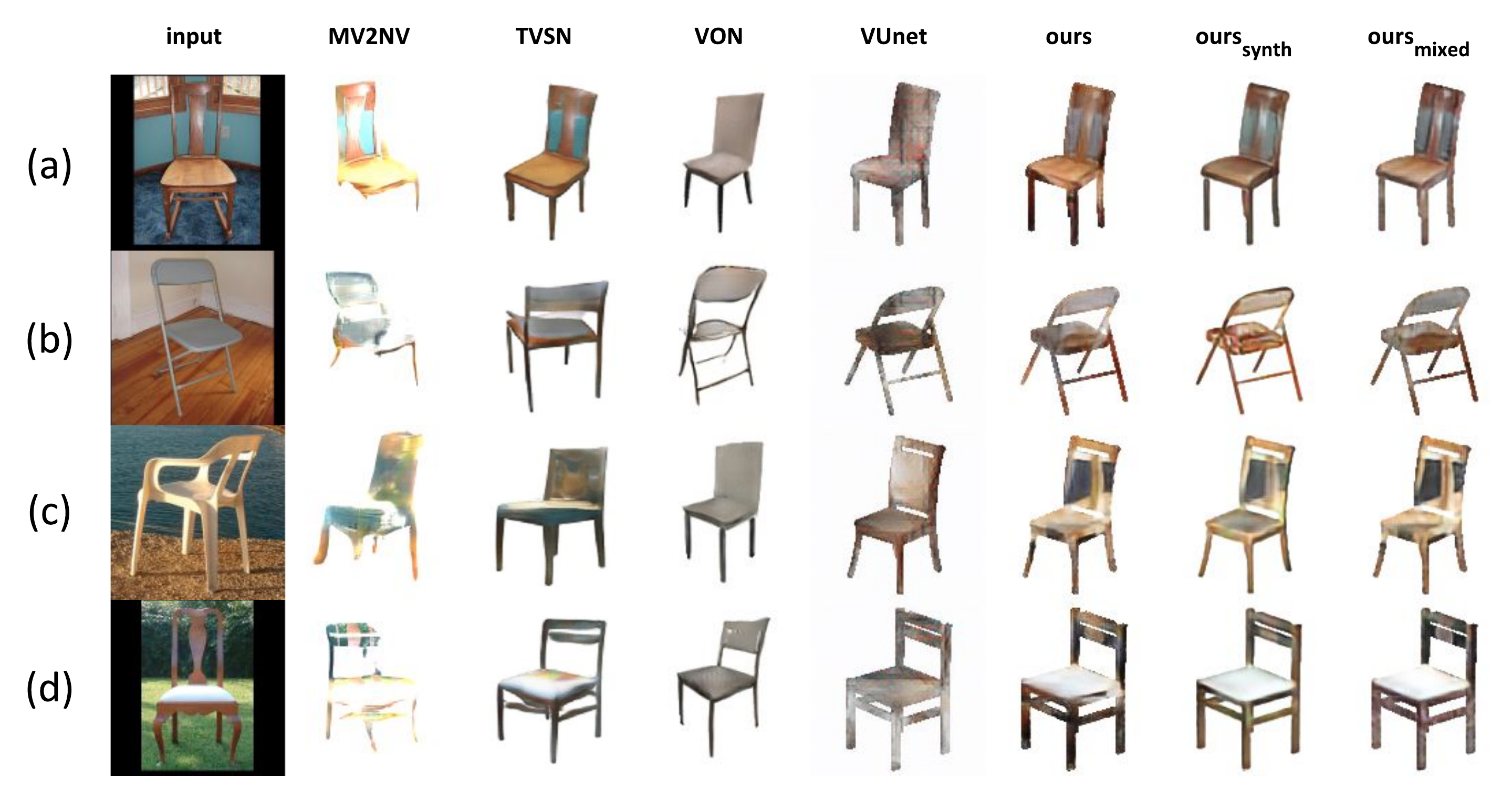}
    \caption{The background leakage is the most common failure case for the \textit{chair} class. This can be solved by providing a 3D model which more closely resembles the one of the input. Please refer to Sec.~\ref{sec:suppl:failure_cases} for details.}
    \label{fig:failure_cases_chairs}
\end{figure*}
We refer the reader to Fig.~\ref{fig:failure_cases_cars} for visual examples of failure cases for the vehicle class. In (a) the \icn is not able to recover in case of input patches which are grossly wrong, either due to a failure in keypoint estimation or to further objects occluding the car such as the people in the first image. Furthermore, we report cases in which the \icn fails to guarantee a realistic output which is consistent with the input image. This happens particularly from rarer viewpoints such as bird's eye views (b) and from the back views (c).\\
Even more than for vehicles, the concave geometry of the \textit{chair} object makes easier for unwanted portions of the background to leak in the generation process. In theory, our method can perfectly deal with these cases, as it masks the generated image with the silhouette of the rendered model. However, in some cases the 3D chair model rendered is so different from the one in the input image that spurious background region make their way to the final output. In Fig.~\ref{fig:failure_cases_chairs} we provide visual examples of this case. Arguably, this problem would be much alleviated if more 3D models were used for inference (at the current state we only choose among the 10 \pascal models).
%
%
%
%
\section{Visualisation Tool}
\begin{figure*}[t]
    \centering
    \includegraphics[width=0.8\textwidth]{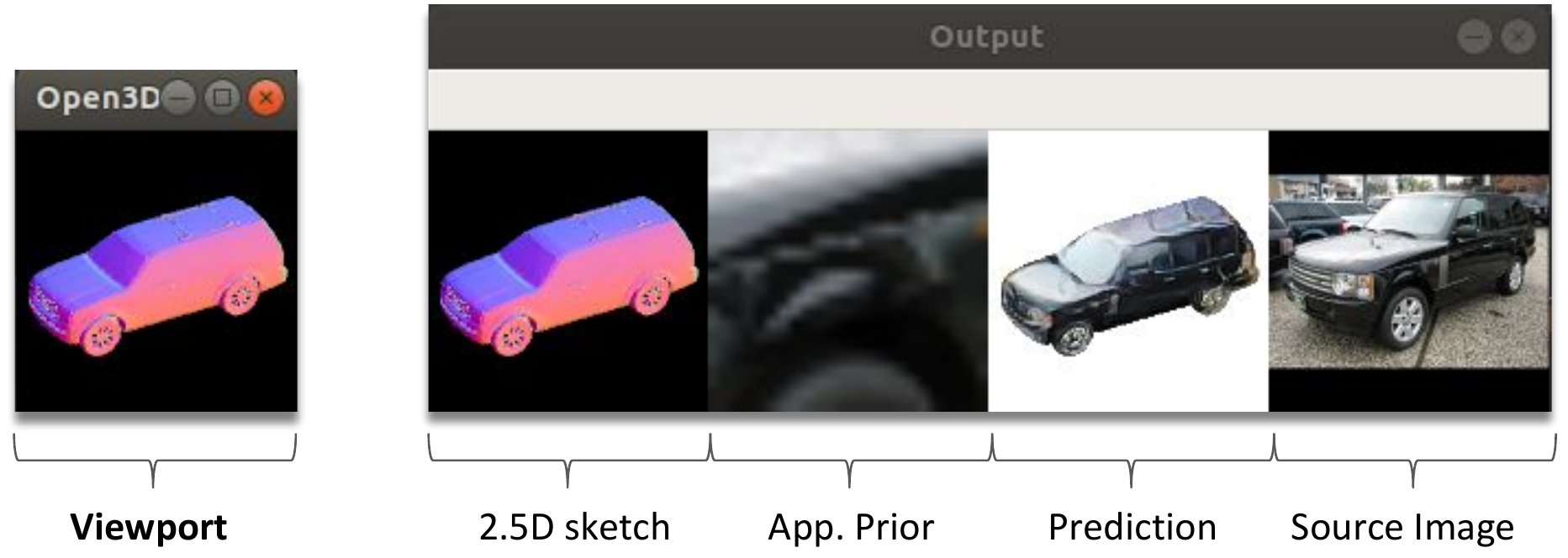}
    \caption{Visualisation Tool interface example.}
    \label{fig:viewport_tool}
\end{figure*}

We release at \href{https://github.com/ndrplz/semiparametric}{https://github.com/ndrplz/semiparametric} the visualization tool we used to inspect the results and create most of the images in this paper. It is written in Python and depends on the Open3D~\cite{zhou2018open3d} and OpenCV~\cite{opencv_library} libraries. The interface simulates a camera moving around an object centered in the origin. The movement are described in terms of spherical coordinates and radius, and each of the three components can be manipulated individually. The user can change:
\begin{itemize}
    \item the elevation value, thus moving from lateral to bird-eye camera;
    \item the azimuth value, thus moving around the object centered in the origin describing a 360 degrees circular trajectory;
    \item the radius value, thus simulating a zoom.
\end{itemize}
The Open3D library provides a 3D environment where the CAD model is loaded, and callbacks enable rendering the scene as a 2D image. Normals visualisation is also supported.
Along with the manipulation of the previous variables, the user can also independently change the CAD model and the appearance image. When started, the tool present the CAD aligned with the input image accordingly to the annotated viewpoint in the data. By changing the spherical coordinates and the radius new viewpoints of the object are synthesised, while by selecting a different CAD model a shape transfer is performed.
An example of the interface is shown in \ref{fig:viewport_tool}.
%
%
%
\begin{figure*}
    \centering
    \includegraphics[width=\textwidth]{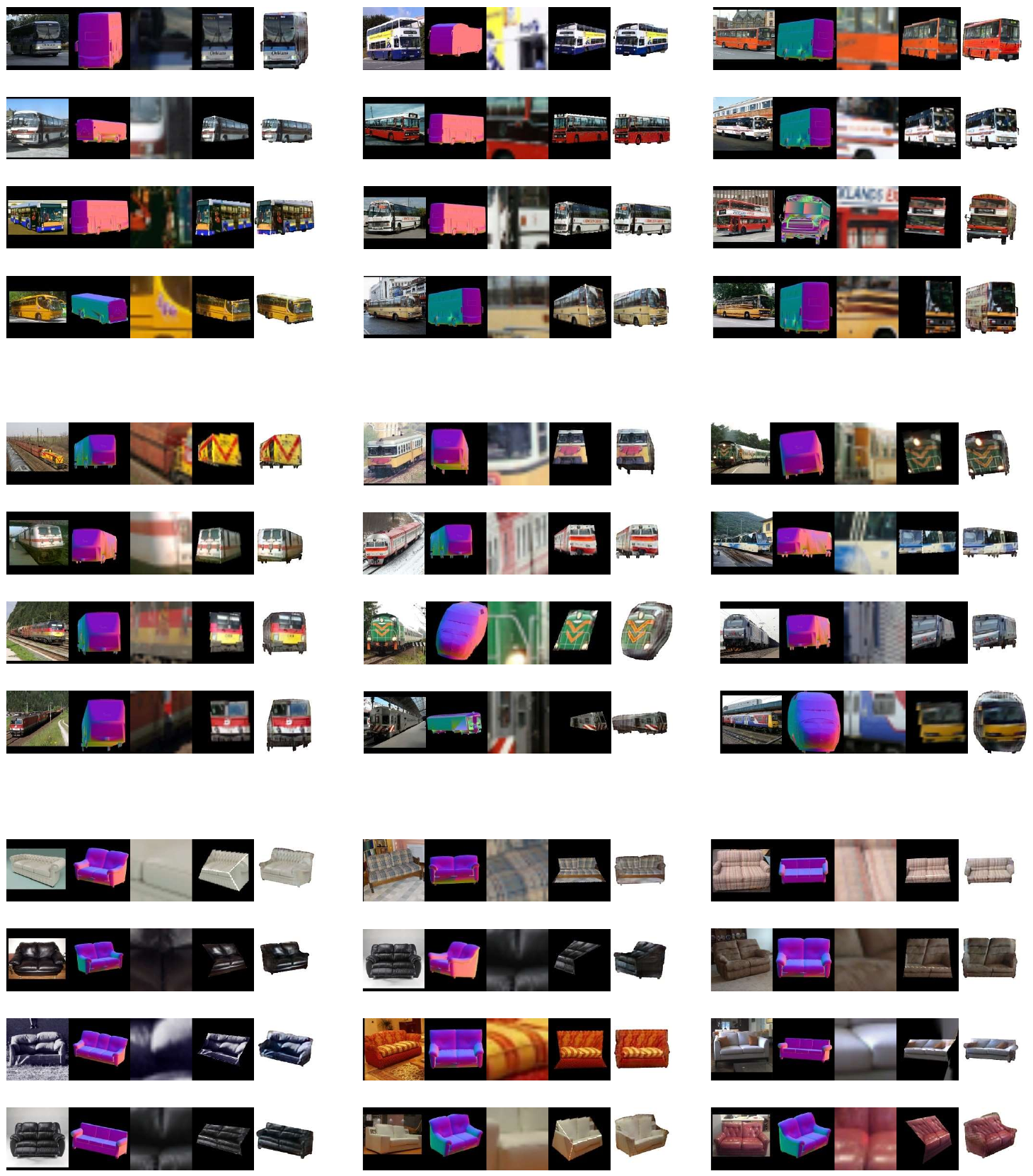}
    \caption{Visual results on additional classes on \pascal dataset. Top to bottom: \textit{Bus}, \textit{Train}, \textit{Sofa}. Left to right for each image: input view, destination 2.5D sketch, appearance prior, warped planes, generated novel view. Best viewed zoomed on screen.}
    \label{fig:suppl:buses_trains_sofas}
\end{figure*}
\section{Visual results on additional classes}
To showcase the generality of our proposed approach, we report in Fig.~\ref{fig:suppl:buses_trains_sofas} visual results on additional classes of \pascal dataset, namely \textit{Bus}, \textit{Train} and \textit{Sofa}. The main challenge for those classes is the much lower number of images compared to the car class. Results show that our framework is able to handle a diverse set of object classes, provided that the underlying constraints (i.e. object symmetry, piece-wise planarity) hold.
%
%
%
\section{Robustness against 2D keypoint errors}
\label{sec:suppl:fid_across_noise}
\begin{figure*}[t]
    \centering
    \includegraphics[width=0.9\textwidth]{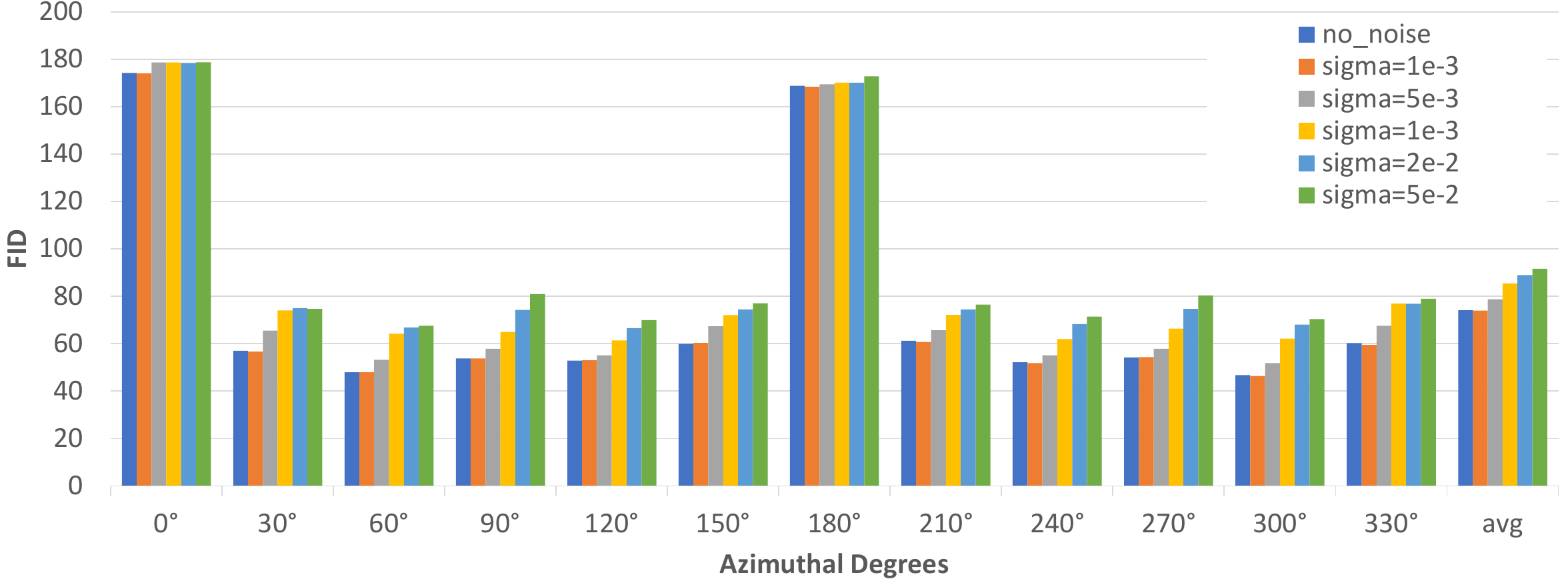}
    \caption{Increase in \fid when adding isotropic Gaussian noise on the 2D keypoints locations. Details in Sec.~\ref{sec:suppl:fid_across_noise}.}
    \label{fig:fid_across_noise}
\end{figure*}
We rely on ground truth keypoints provided in the \pascal dataset for training and evaluation. Still, to assess how the model would react to keypoints predicted \enquote{in-the-wild}, we experiment with adding isotropic Gaussian noise to the 2D keypoints locations at evaluation time. Fig.\ref{fig:fid_across_noise} shows how the \fid is affected by noises with increasing variance (it is worth noting that keypoints are normalized in range $[0,1]$). Although the results obviously degrade with respect to the annotated keypoints locations, we notice that overall performance is still competitive against most competitors even when adding aggressive noises.
%
%
\section{Additional ablation: only appearance prior}
\label{sec:suppl:ablation_only_prior}
\begin{figure*}
    \centering
    \includegraphics[width=\textwidth]{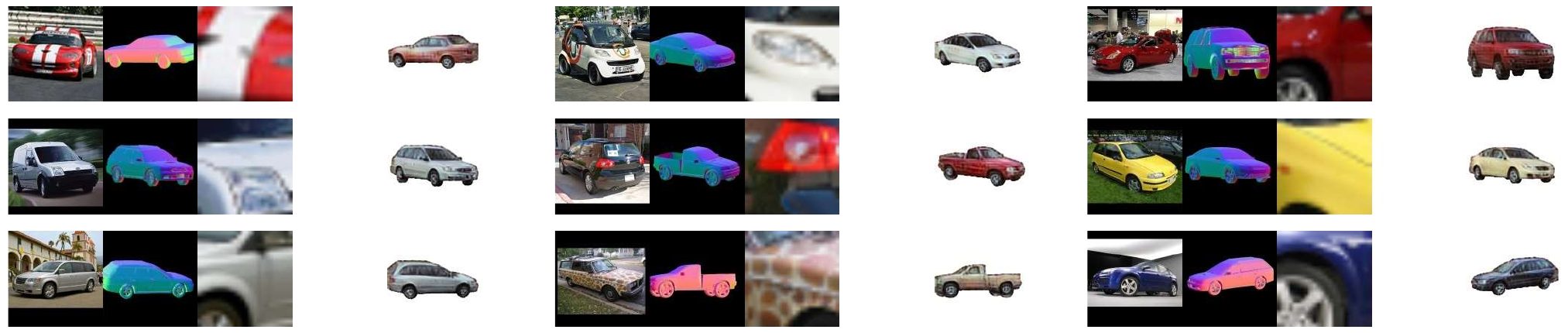}
    \caption{Visual results from appearance-only ablation study. Here, the model can access only the 2.5D sketch and the central crop from the input view. Details in Sec.~\ref{fig:suppl:ablation_only_prior}.}
    \label{fig:suppl:ablation_only_prior}
\end{figure*}
Figure~\ref{fig:suppl:ablation_only_prior} depicts visual results for a network trained using the appearance prior only. While the general color from the central crop is applied to the destination image, almost all other details are lacking. This emphasises the importance of the image planes as the main source of texture in the final image. The \fid score for this ablated version (averaged across all 12 angles) is 154.2. 
%
%
%
\section{A/B Perceptual Experiments}
\begin{figure*}[t]
\centering
\begin{tabular}{c}
    \includegraphics[width=0.95\textwidth]{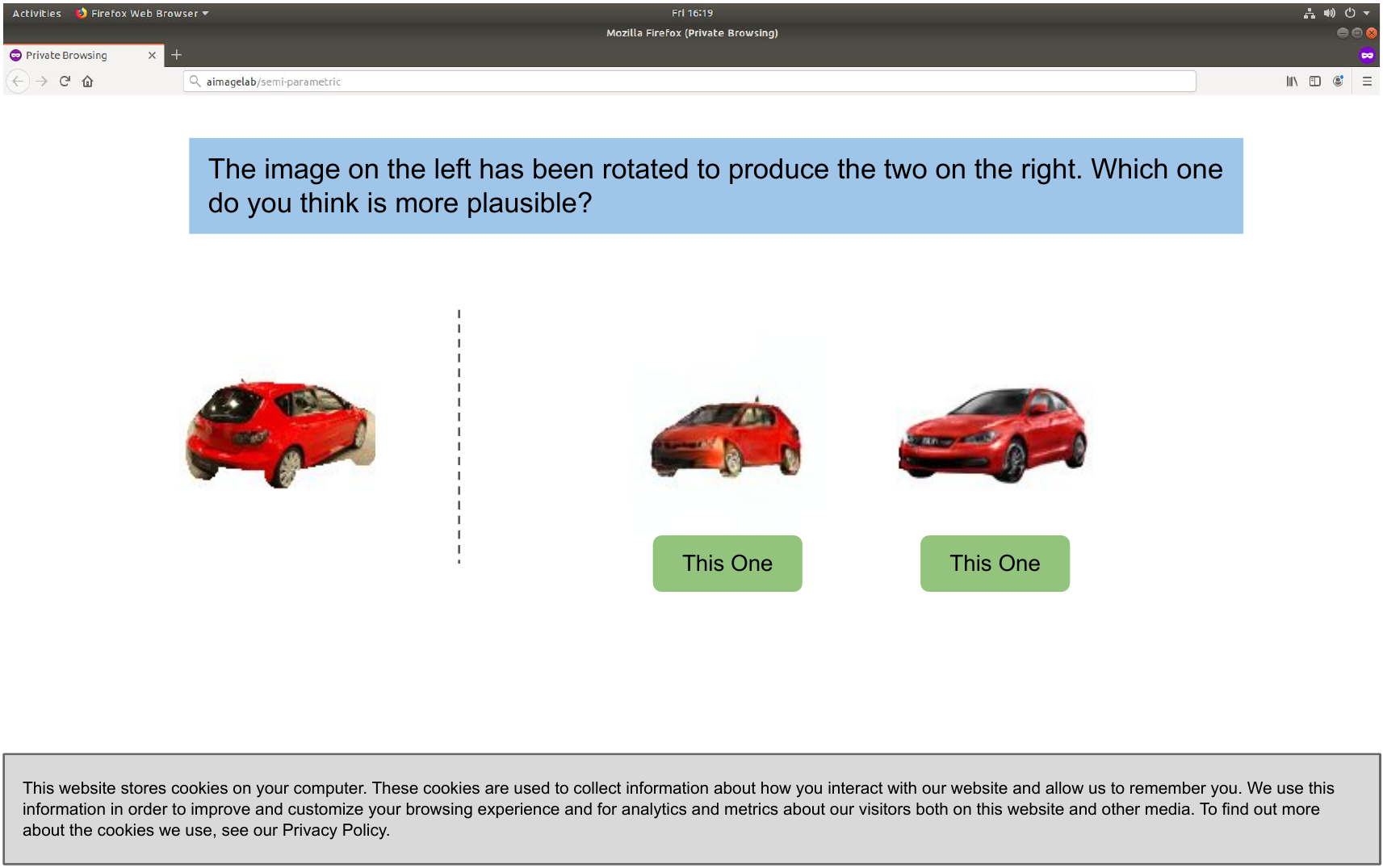} \\
    \includegraphics[width=0.95\textwidth]{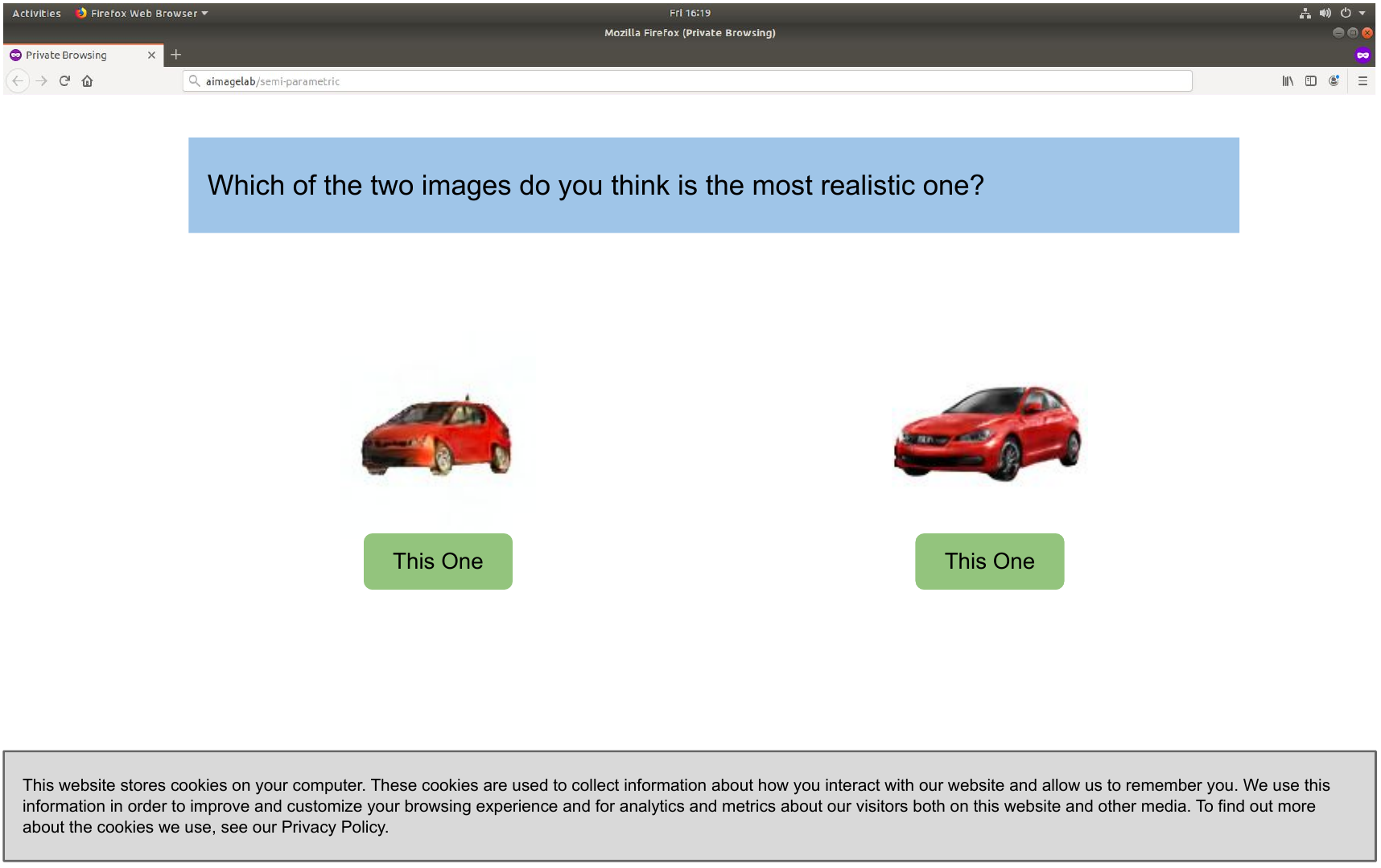}
\end{tabular}
    \label{fig:ab}
    \caption{Screenshots for the two A/B preference tests settings.}
\end{figure*}
We developed a tool to perform A/B test preferences based on Django web server. The system is centralised and can be accessed from multiple terminals simultaneously. We wrote asking for volunteers on the University of Modena and Reggio Emilia mailing list, and the tests were performed from the volunteer's computer. We didn't interact directly with participants to avoid introducing any bias inadvertently. To ensure the correctness of the data we discarded the first results from each session, which were considered as a briefing.
For all experiments, images were shown at the same resolution of 128x128 pixels. As all methods produce a white background, ground truth aligned 3D CAD is used to mask \pascal real images. Both sampling order and left-right order of \textit{A} and \textit{B} were randomized.
Raw results have been stored into a MySQL DB to ensure persistence, while aggregated statistics were presented on an administrator interface.
Fig.~16 depicts screenshots from the two perceptual experiments described in Sec.4 of the main paper.
%
%
%
%
\ifCLASSOPTIONcaptionsoff
  \newpage
\fi
\end{document}